%% file: main.tex
\documentclass[10pt,twocolumn,letterpaper]{article}

\usepackage{euvs}              

\input{preamble}

\definecolor{euvsblue}{rgb}{0.21,0.49,0.74}
\definecolor{nvgreen}{rgb}{0.47,0.73,0}
\usepackage[pagebackref,breaklinks,colorlinks,allcolors=euvsblue]{hyperref}
\usepackage{float}
\usepackage{ulem}
\usepackage{multicol,multirow}
\usepackage{soul}
\usepackage{xcolor} 
\usepackage{colortbl}
\sethlcolor{orange!30}
\ifdefined\todo
\else
  \usepackage{todonotes}
\fi

\def\paperID{6470} 
\def\confName{euvs}
\def\confYear{2025}

\title{Extrapolated Urban View Synthesis Benchmark}

\author{
  Xiangyu Han\textsuperscript{1,3}\thanks{Equal contribution.} \quad
  Zhen Jia\textsuperscript{1}\footnotemark[1] \quad
  Boyi Li\textsuperscript{2} \quad
  Yan Wang\textsuperscript{2} \quad
  Boris Ivanovic\textsuperscript{2} \quad
  Yurong You\textsuperscript{2} \\
  Lingjie Liu\textsuperscript{3} \quad
  Yue Wang\textsuperscript{2,4} \quad
  Marco Pavone\textsuperscript{2,5} \quad
  Chen Feng\textsuperscript{1} \quad
  Yiming Li\textsuperscript{1,2}\thanks{Corresponding author.}\footnotemark[2] \vspace{2mm} \\ 
  $^{1}$NYU \quad 
  $^{2}$NVIDIA \quad 
  $^{3}$UPenn \quad 
  $^{4}$USC \quad 
  $^{5}$Stanford \\
  {\small \url{https://ai4ce.github.io/EUVS-Benchmark}}
}

\begin{document}
\twocolumn[{
    \renewcommand\twocolumn[1][]{#1}
    \maketitle
    \vspace{-2em}
    \begin{center}
        \centering
        \includegraphics[width=0.99\textwidth]{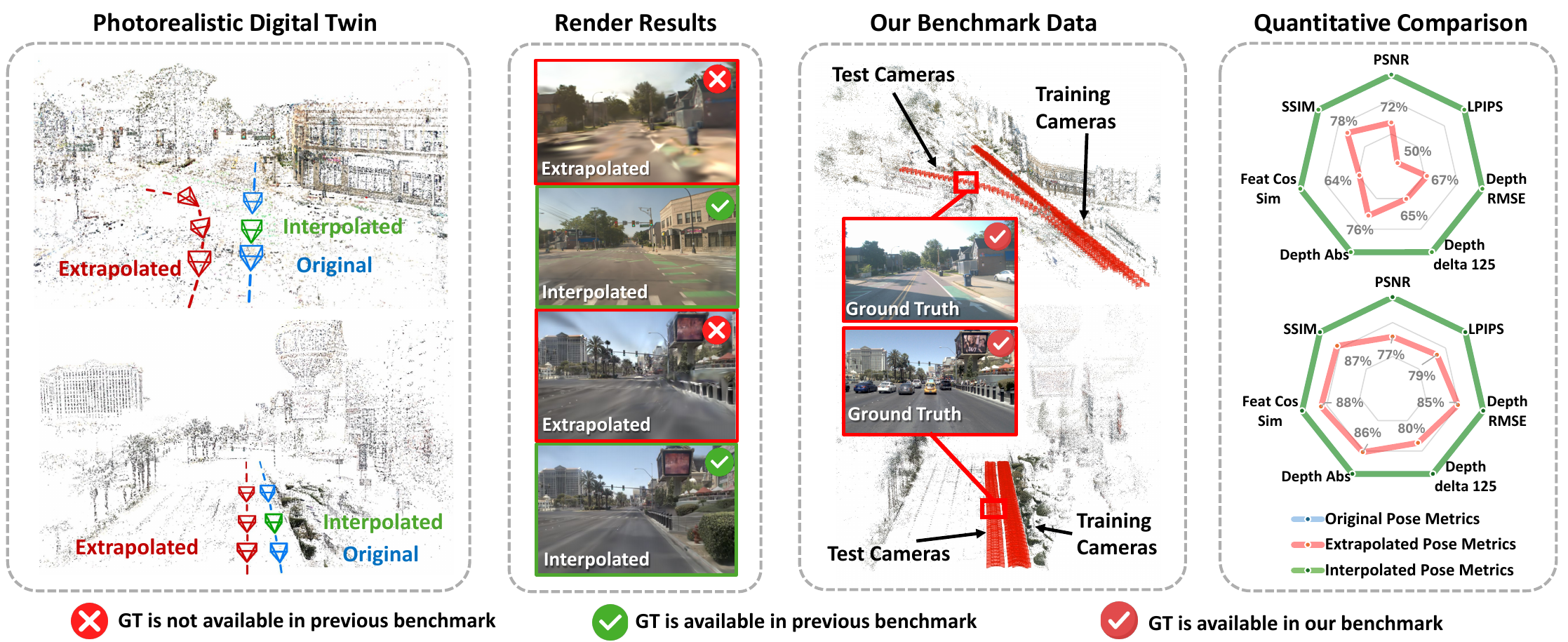}
        \captionof{figure}{\textbf{Our key contributions.} Previous evaluations for urban view synthesis have primarily focused on interpolated poses, as the lack of ground truth data has made it challenging to evaluate extrapolated poses. We address this gap by providing real-world data that enables both quantitative and qualitative evaluations of extrapolated view synthesis in urban scenes. The quantitative results reveal a significant performance drop in 3D Gaussian Splatting~\cite{kerbl3Dgaussians} when handling extrapolated views, highlighting the need for more robust NVS methods.
        }
        \label{fig:teaser}
    \end{center}
}]

\footnotetext[1]{Equal contribution.} 
\footnotetext[2]{Corresponding author.}

\input{sec/0_abstract}    
\input{sec/1_intro}
\input{sec/2_related}

\input{sec/3_EUVS_benchmark}

\input{sec/4_experiment}

\input{sec/5_discussion}
\input{sec/6_conclusion}

\clearpage
\clearpage

{
    \small
    \bibliographystyle{ieeenat_fullname}
    \bibliography{main}
}

\input{sec/X_supplementary}

\end{document}

%% file: preamble.tex
\newcommand{\todo}[1]{{\color{red}#1}}


%% file: sec/0_abstract.tex
\begin{abstract}
Photorealistic simulators are essential for the training and evaluation of vision-centric autonomous vehicles (AVs). At their core is Novel View Synthesis (NVS), a crucial capability that generates diverse unseen viewpoints to accommodate the broad and continuous pose distribution of AVs. Recent advances in radiance fields, such as 3D Gaussian Splatting, achieve photorealistic rendering at real-time speeds and have been widely used in modeling large-scale driving scenes. However, their performance is commonly evaluated using an interpolated setup with highly correlated training and test views. In contrast, extrapolation, where test views largely deviate from training views, remains underexplored, limiting progress in generalizable simulation technology. To address this gap, we leverage publicly available AV datasets with multiple traversals, multiple vehicles, and multiple cameras to build the first \textbf{E}xtrapolated \textbf{U}rban \textbf{V}iew \textbf{S}ynthesis (EUVS) benchmark. Meanwhile, we conduct both \textbf{quantitative} and \textbf{qualitative} evaluations of state-of-the-art NVS methods across different evaluation settings. Our results show that current NVS methods are prone to overfitting to training views. Besides, incorporating diffusion priors and improving geometry cannot fundamentally improve NVS under large view changes, highlighting the need for more robust approaches and large-scale training. We will release the data to help advance self-driving and urban robotics simulation technology.

\end{abstract}

%% file: sec/1_intro.tex
\section{Introduction}
\label{sec:intro}
The development of vision-centric autonomous vehicles (AVs) relies heavily on photorealistic simulators, which provide controlled, reproducible, and scalable environments for training and evaluation of driving models~\cite{dosovitskiy2017carla,yang2023unisim,tonderski2024neurad}. These simulators enable AVs to learn and adapt to a variety of real-world scenarios, from crowded urban streets to adverse weather conditions, without the logistical and safety concerns of physical road testing. At the heart of these simulators is the capability for Novel View Synthesis (NVS)—a key technology that generates realistic images of unseen viewpoints, simulating the continuous changes in perspective that occur as AVs navigate through urban environments.

Recent advancements in radiance fields, particularly methods based on 3D Gaussian Splatting~\cite{kerbl3Dgaussians}, have significantly improved the realism and efficiency of NVS. These approaches~\cite{chen2023periodic,zhou2024drivinggaussian,yan2024street,zhou2024hugs} can produce photorealistic renderings at real-time speeds, making them highly attractive for large-scale driving scene simulation. However, despite their impressive results, the evaluation of NVS methods has predominantly focused on \textbf{interpolated} scenarios, where training and test viewpoints are closely related. While interpolation tests are valuable for assessing local consistency, they fall short in addressing the more critical challenge of \textbf{extrapolation}—where test viewpoints differ significantly from the training data. As shown in \autoref{fig:teaser}, the interpolation test set demonstrates strong performance, with metrics such as PSNR, SSIM, and LPIPS remaining very close to the training set values. In contrast, the extrapolation test set, which includes additional translation and rotation changes relative to the training set, exhibits notable drops in performance. Specifically, the metric decreases relative to the training set are \textbf{28\%} for PSNR, \textbf{22\%} for SSIM, and \textbf{50\%} for LPIPS. These results underscore the urgent need to explore and advance extrapolated view synthesis in complex urban scenes, as real-world driving often involves encountering scenarios with significant viewpoint shifts and diverse spatial transformations that deviate from training distributions. Several recent studies~\cite{hwang2024vegs,han2024ggs} have investigated the generalization capabilities of NVS in 3D Gaussian Splatting. Although they show promising qualitative results, there is \textbf{no comprehensive quantitative analysis} due to the absence of standardized datasets. Moreover, their evaluations are primarily limited to specific scenarios or use cases, without investigating varying evaluation settings based on the degree of extrapolation. This gap underscores the urgent need for a benchmark that offers diverse and challenging datasets, enabling a rigorous and systematic assessment of NVS methods.

To establish a common platform for assessing the robustness of NVS methods, we introduce a comprehensive benchmark for quantitatively and qualitatively evaluating extrapolated novel view synthesis in large-scale urban scenes. Our benchmark leverages publicly available datasets, including NuPlan~\cite{caesar2021nuplan}, MARS~\cite{li2024multiagent}, and Argoverse2~\cite{wilson2023argoverse}, which feature multi-traversal, multi-agent and multi-camera sensory recordings. Multi-traversal data consists of asynchronous traversals of the same location, while multi-agent data refers to data collected simultaneously from multiple vehicles within the same area. These data provide diverse camera poses within a 3D scene, enabling the training and evaluation of extrapolated view synthesis in outdoor environments. For the experimental setup, we define three evaluation settings: (1) translation only, (2) rotation only, and (3) translation + rotation, as shown in~\autoref{fig:EUVS benchmark figure}. In autonomous driving scenarios, Setting 1 corresponds to maneuvers such as lane changes, Setting 2 involves switching between cameras facing different directions, and Setting 3 addresses complex intersections, such as crossroads with diverse traversal paths. These settings represent common challenges in autonomous driving, and addressing them enables the synthesis of complete scenes from sparse image observations.

We conduct pose estimation and sparse reconstruction using COLMAP~\cite{schoenberger2016sfm}, which facilitates the initialization of Gaussian Splatting. We then evaluate state-of-the-art Gaussian Splatting-based approaches across each evaluation setting, identifying performance gaps both qualitatively and quantitatively in extrapolated urban view synthesis.

In summary, our main contributions are as follows:
\begin{itemize}
    \item We initiate the first comprehensive quantitative and qualitative study on the Extrapolated Urban View Synthesis (EUVS) problem, supported by a robust evaluation framework that categorizes evaluation settings (translation-only, rotation-only, and translation + rotation) while assessing performance using diverse metrics, including reconstruction accuracy and visual fidelity.
    \item We construct a novel dataset by integrating multi-traversal, multi-agent, and multi-camera data from publicly available resources, totaling \textbf{90,810} frames across \textbf{345} videos. Our dataset effectively addresses the limitations of existing benchmarks, enabling rigorous and robust evaluation for extrapolated urban view synthesis.
    \item We benchmark state-of-the-art Gaussian Splatting-based and NeRF-based models and analyze key factors that influence the performance of extrapolated NVS, laying a solid foundation for future advancements in this challenging task. Data and code will be released upon acceptance.
\end{itemize}

%% file: sec/2_related.tex
\section{Related Works}

\noindent \textbf{Extrapolated View Synthesis.}
Extrapolated view synthesis aims to generate novel views beyond observed perspectives, addressing challenges in visual coherence for unseen regions. RapNeRF~\cite{zhang2022ray} proposes a random ray-casting policy that enables training on unseen views based on visible ones. Following work~\cite{yang2023nerfvs} enhances this approach by incorporating holistic priors. Besides, some generalizable models~\cite{wewer2024latentsplat,chen2024mvsplat,charatan2024pixelsplat} have emerged, capable of generating extrapolated novel views from a limited number of input images. While these methods are designed for indoor scenes, several works address extrapolated view synthesis in outdoor driving scenarios, which typically involve forward-facing cameras and unbounded environments. 
To tackle the Setting 1 challenge in our benchmark and address the scarcity of lane change data, GGS~\cite{han2024ggs} introduces a novel virtual lane generation module. In parallel, AutoSplat~\cite{khan2024autosplat} tackles lane change in dynamic scenes by applying geometric and reflected consistency constraints. To address the Setting 1 or 2 challenge, FreeSim~\cite{fan2024freesimfreeviewpointcamerasimulation}, VEGS~\cite{hwang2024vegs}, and SGD~\cite{yu2024sgd}  enhance 3DGS~\cite{kerbl3Dgaussians} with diffusion priors to improve generalization ability. \textit{Yet existing methods suffer from two major limitations: (1) a lack of real data for quantitative evaluation, which confines them to quantitative analysis, and (2) a narrow focus on a specific setting in our benchmark, preventing a comprehensive and systematic exploration.}

\noindent \textbf{3D Gaussian Splatting.}
Recent advances in radiance fields, particularly NeRF~\cite{mildenhall2021nerf} and 3DGS~\cite{kerbl3Dgaussians}, have garnered significant attention due to their impressive advancements in NVS. NeRF employs an implicit representation through a multi-layer perceptron (MLP). Furthermore, 3DGS explicitly represents scenes using anisotropic 3D Gaussian ellipsoids, enabling high-quality real-time rendering. Several works have addressed issues such as difficulties with reflective surfaces~\cite{jiang2024gaussianshader}, aliasing~\cite{yu2024mip}, etc. However, urban scenes introduce unique challenges due to their unbounded and dynamic nature. To address the challenge, several works separate dynamic and static elements in the scene by leveraging a composite dynamic Gaussian graph~\cite{zhou2024drivinggaussian,yan2024street}, optical flow prediction~\cite{zhou2024hugs,yang2023emernerf}, etc. PVG~\cite{chen2023periodic} presents a unified representation model that simultaneously incorporates both dynamic and static components without relying on priors. To achieve realistic geometry and efficient rendering, 2DGS~\cite{huang20242d} collapses 3D Gaussians onto 2D planes, while hybrid approaches~\cite{wu2024hgs,shi2024dhgs} combine different Gaussians to better capture region-specific features. \textit{In summary, current urban NVS methods primarily focus on effectively handling dynamic elements and enhancing geometry representation, while the challenge of extrapolated view synthesis remains largely underexplored.}

\noindent \textbf{Autonomous Driving Simulators.}
Current simulators focus on three key challenges: parameter initialization~\cite{tan2021scenegen,feng2023trafficgen}, traffic simulation~\cite{zhong2023guided,xu2023bits,li2024scenarionet}, and sensor simulation~\cite{amini2022vista,gao2024vista,kim2021drivegan,zhao2024drivedreamer4d,yang2023unisim,dosovitskiy2017carla}. Sensor simulation is crucial for generating realistic sensory data that AVs depend on for perception and decision-making. Early sensor simulators~\cite{dosovitskiy2017carla,shah2018airsim,wymann2000torcs} provide simulated environments that are valuable for research but lack visual realism. Recent studies have focused on data-driven simulators that extract data from real-world driving logs, creating more realistic and adaptable environments. These methods can be classified into two categories: generation-based and reconstruction-based approaches. The former rely on inputs such as text, video, and other data sources for simulation, supported by world models or prior knowledge~\cite{kim2021drivegan,zhao2024drivedreamer4d,gao2024vista}. Reconstruction-based simulations leverage real-world data to ensure both visual fidelity and geometric consistency~\cite{wu2023mars,yang2023unisim,wu2023mapnerf}. UniSim~\cite{yang2023unisim} is a pioneering example of this approach, utilizing NeRF-based scene representation to create dynamic scenes with geometric information that are both editable and controllable. \textit{Extrapolated view synthesis is essential for these simulators, as it enables the generation of realistic and consistent views from diverse angles.}

\noindent \textbf{Autonomous Driving Datasets.} 
High-quality datasets play a vital role in advancing autonomous driving research. The KITTI dataset~\cite{geiger2012we}, released in 2012, marked a major milestone, significantly accelerating advances in AVs~\cite{teeti2023vision,fritsch2013new,menze2015object}. Since then, many influential autonomous vehicle datasets have been developed to tackle challenges like adverse weather conditions~\cite{pitropov2021canadian}, multimodal fusion~\cite{caesar2020nuscenes,caesar2021nuplan}, repeated driving~\cite{diaz2022ithaca365,caesar2021nuplan}, collaborative driving~\cite{xu2023v2v4real,li2022v2x,li2024multiagent}, and motion prediction~\cite{ettinger2021large,chang2019argoverse,wilson2023argoverse}, etc. We leverage publicly available datasets with multi-traversal, multi-agent, and multi-camera recordings, enabling a comprehensive and robust evaluation of extrapolated urban view synthesis.

%% file: sec/3_EUVS_benchmark.tex
\begin{figure*}[ht]
    \centering
    \includegraphics[width=\textwidth]{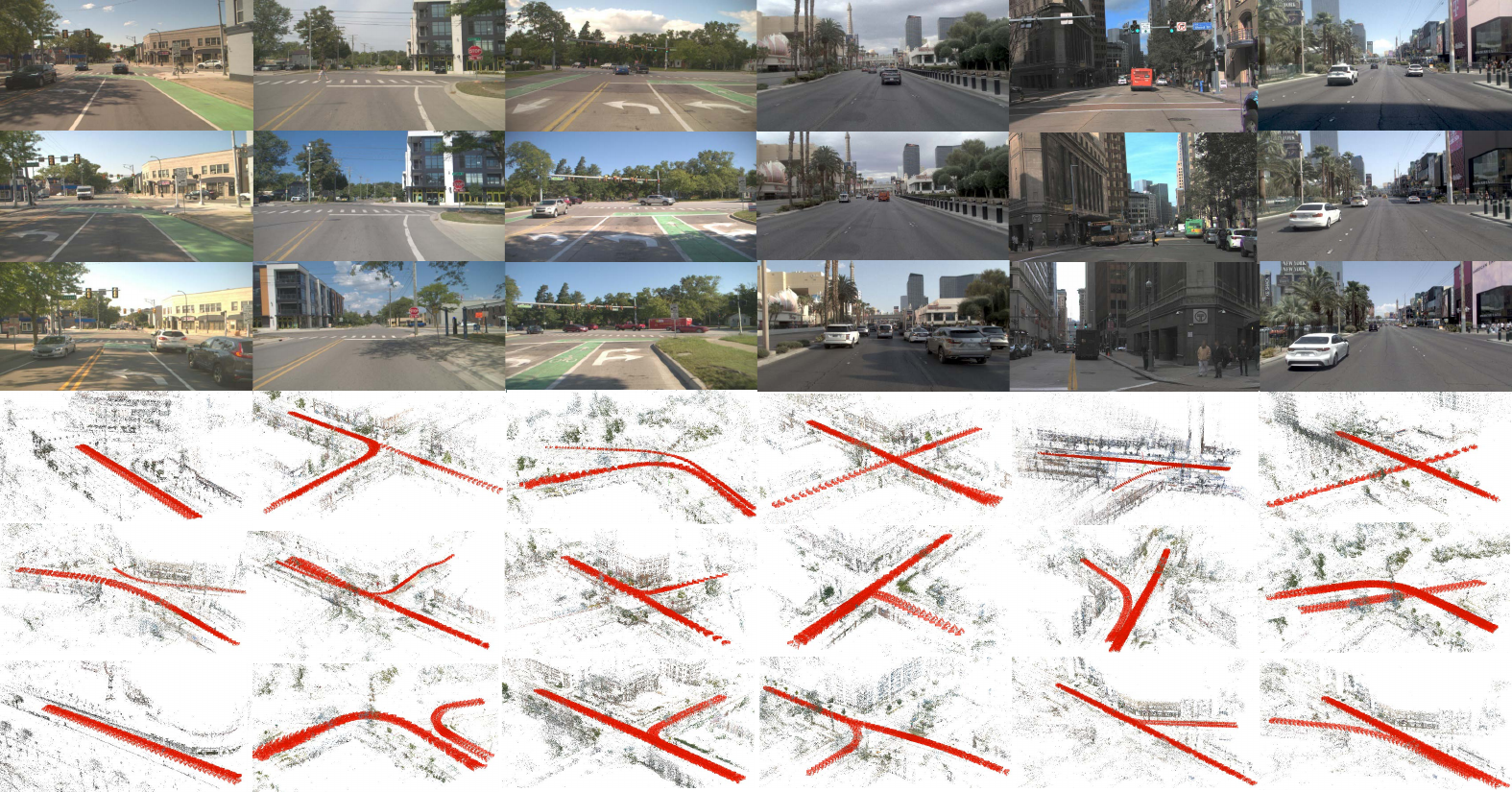}
    \caption{\textbf{Dataset visualization.} Our dataset features diverse scenes across various locations in different cities, sourced from multiple datasets. Typical driving scenarios include maneuvers such as lane changes, cross intersections, and T-junctions. \textbf{Top:} Each column displays images captured at the same location by different agents or traversals. \textbf{Bottom:} Each image displays the COLMAP points at a specific location, along with the corresponding camera poses.}
    \label{fig:dataset overview}
    \vspace{-1.5mm}
\end{figure*}


\begin{figure*}[ht]
    \centering
    \includegraphics[width=\textwidth]{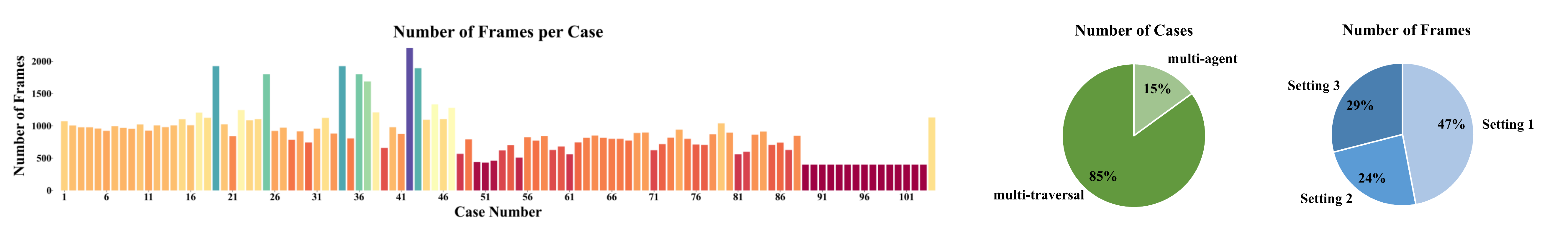}

    \caption{\textbf{Dataset distribution.} Our dataset comprises \textbf{90,810} frames distributed over \textbf{104} cases, capturing a diverse array of multi-traversal paths, multi-agent interactions, and multi-camera perspectives across varying evaluation settings.}
    \label{fig:dataset_distribution}
    \vspace{-2mm}
\end{figure*}

\section{The EUVS Benchmark}

\subsection{Dataset Curation}
We leverage three publicly available, community-verified autonomous driving datasets—nuPlan~\cite{caesar2021nuplan}, Argoverse 2~\cite{wilson2023argoverse}, and MARS~\cite{li2024multiagent}—to enhance adoption and foster trust through their established reliability.
nuPlan~\cite{caesar2021nuplan} provides 1,200 hours of driving data from four cities, serving as the first large-scale planning benchmark. Argoverse 2~\cite{wilson2023argoverse} supports multimodal perception and forecasting with 1,000 annotated 3D scenarios, 20,000 unlabeled lidar sequences, and 250,000 motion forecasting cases. MARS~\cite{li2024multiagent} enables collaborative driving research with multi-agent and multi-traversal scenarios.
The original datasets were not designed for evaluating extrapolated view synthesis, which requires our significant labor in data processing—\textbf{300+ hours} of manual traversal selection and \textbf{800+ hours} of computing to run COLMAP.
By integrating these datasets, our benchmark enables the evaluation of view synthesis across diverse and realistic urban environments under varying conditions.
\autoref{fig:dataset_distribution} illustrates the distribution of the integrated datasets.

\subsection{Evaluation Framework}
\label{sec:evaluation framework}
\begin{figure*}[ht]
    \centering
    \includegraphics[width=0.99\textwidth]{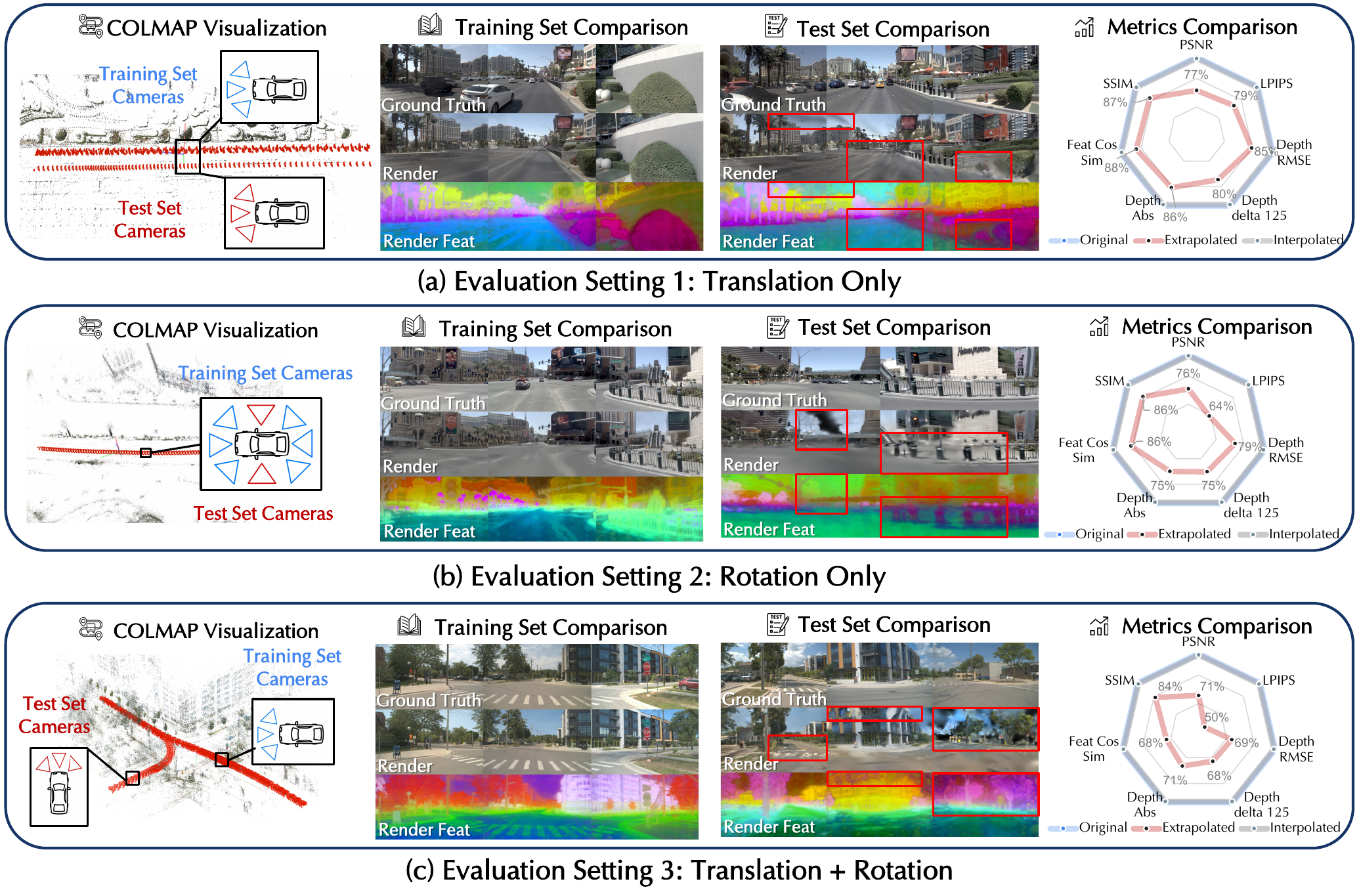}
    \caption{\textbf{Qualitative and quantitative results across three evaluation settings.} The performance drop from interpolation to extrapolation is significant in both qualitative and quantitative comparison. Different testing settings have distinct scenario characteristics, enabling the evaluation of a method's capabilities from various aspects, thus systematically assessing the overall performance of reconstruction algorithms, including geometric accuracy, hallucination ability, view consistency, and depth precision, etc.}
    \label{fig:EUVS benchmark figure}
    \vspace{-2.5mm}
\end{figure*}

To systematically assess model performance in extrapolated urban view synthesis, our evaluation framework incorporates \textit{three evaluation settings} and \textit{three data configurations}. Data configurations include multi-traversal, multi-agent, and multi-camera, while evaluation settings are categorized into (1) translation only, (2) rotation only, and (3) translation + rotation, as illustrated in ~\autoref{fig:EUVS benchmark figure}.

\noindent \textbf{Setting 1.} The translation-only experimental setup involves scenarios where the vehicle's position shifts without any change in orientation. This scenario is commonly observed in lane changes. We use traversals from different lanes in multi-traversal data, focusing on the three front cameras. The data is sourced from nuPlan~\cite{caesar2021nuplan} and Argoverse~2~\cite{wilson2023argoverse}.

\noindent \textbf{Setting 2.} The second setting, rotation only, evaluates models on views with significant orientation changes. In vision-centric autonomous vehicles, this corresponds to transitions between cameras capturing different directions. We leverage multi-camera data from nuPlan~\cite{caesar2021nuplan}, training on three forward-facing and three rear-facing cameras, while evaluating on two side-facing cameras.

\noindent \textbf{Setting 3.} The third setting, combining translation and rotation, includes both positional shifts and orientation changes, posing the greatest challenge for NVS. To address this, we utilize multi-traversal driving data collected from the same location but across different traversal routes. For example, the training and test sets may include routes that approach an intersection from different directions. Typical route combinations feature scenarios such as intersections, T-junctions, and Y-junctions, as shown in \autoref{fig:dataset overview}, ensuring diversity and comprehensive evaluation. The data for this setting is from MARS~\cite{li2024multiagent} and Argoverse 2~\cite{wilson2023argoverse}.

\subsection{Algorithm Overview}
\noindent \textbf{Vanilla 3D Gaussian Splatting.} 3D Gaussian Splatting (3DGS)~\cite{kerbl3Dgaussians} leverage 3D Gaussians to explicitly represent the scene, which achieves high quality while offering real-time rendering by avoiding unnecessary computation in the empty space. Building on this, 3DGM~\cite{li2024memorize} leverages multi-traversal consensus to differentiate transient and permanent elements, enabling joint 2D segmentation and 3D mapping without using any human supervision.

\noindent \textbf{Planar-based and Geometry Refined Gaussian Splatting.}
GaussianPro~\cite{cheng2024gaussianpro} builds on 3DGS~\cite{kerbl3Dgaussians} by introducing multi-frame geometric optimization, which guides the densification of 3D Gaussians, enhancing scene consistency in complex geometries. It further refines geometry by encouraging Gaussian primitives to adopt flat structures. Similarly, 2DGS~\cite{huang20242d} projects the 3D volume into a set of 2D oriented planar Gaussian disks, enabling high-fidelity surface reconstruction. PGSR~\cite{chen2024pgsr} introduces an unbiased depth rendering method and integrates single-view geometric, multi-view photometric, and geometric regularization techniques to improve global geometry accuracy.

\noindent \textbf{Gaussian Splatting with Diffusion Priors.}
VEGS~\cite{hwang2024vegs} introduces a novel view generalization approach that harnesses pre-extracted surface normals to align 3D Gaussians while generating augmented camera views guided by diffusion priors. These diffusion priors serve a dual purpose: providing denoising loss guidance and supervising the training of augmented cameras. This process effectively mitigates floating artifacts and fragmented geometries, resulting in more accurate and coherent 3D representations.

\noindent \textbf{Feature-Enhanced Gaussian Splatting.} Feature 3DGS~\cite{zhou2024feature} extends 3D Gaussian Splatting with a Parallel N-dimensional Gaussian Rasterizer, allowing simultaneous rendering of radiance fields and high-dimensional semantic features. By embedding semantic features directly into 3D Gaussians, the approach enhances optimization, enabling better correspondence with scene semantics and achieving more detailed and accurate spatial representations.

\noindent \textbf{NeRF-based Method.}
Instant-NGP~\cite{mueller2022instant} uses a multiresolution hash encoding to map spatial coordinates into compact latent representations via hash tables. It efficiently encodes high-frequency details by combining trainable feature vectors with interpolation, enabling adaptive and scalable input encodings. 
Zip-NeRF~\cite{barron2023zip} leverages multisampling with isotropic Gaussians for scale-aware features and introduces a smooth anti-aliasing loss to address z-aliasing. 
In addition, it incorporates a novel distance normalization technique to better manage close and distant objects, achieving high-quality rendering and fast training.

%% file: sec/4_experiment.tex
\sethlcolor{orange!40}
\definecolor{orange20}{HTML}{FFECD6} 
\definecolor{orange30}{HTML}{FFD4A3} 
\definecolor{orange40}{HTML}{FFBB77} 
\definecolor{orange50}{HTML}{FFAD33}

\begin{table*}[t]
\footnotesize
    \centering
    \caption{\textbf{Quantitative rendering results across three evaluation settings.} \textit{In.} denotes interpolation, while \textit{Ex.} represents extrapolation. Different baselines excel in different settings, reflecting the comprehensiveness and completeness of the evaluation protocol.}
    \setlength{\tabcolsep}{1.8mm}{
    \begin{tabular}{llcccccccccccc}
        \toprule
        & \multirow{2}{*}{\textbf{Method}} & \multicolumn{3}{c}{\textbf{PSNR}\,$\uparrow$} & \multicolumn{3}{c}{\textbf{SSIM}\,$\uparrow$} & \multicolumn{3}{c}{\textbf{LPIPS}\,$\downarrow$} & \multicolumn{3}{c}{\textbf{Feat Cos Sim}\,$\uparrow$} \\
        \cmidrule(lr){3-5} \cmidrule(lr){6-8} \cmidrule(lr){9-11} \cmidrule(lr){12-14}
        & & \textbf{In.} & \textbf{Ex.} & \textbf{Drop} & \textbf{In.} & \textbf{Ex.} & \textbf{Drop} & \textbf{In.} & \textbf{Ex.} & \textbf{Drop} & \textbf{In.} & \textbf{Ex.}  & \textbf{Drop}\\
        \midrule
        \multirow{9}{*}{\textbf{Setting 1}} 
        & 3DGS~\cite{kerbl3Dgaussians} & 21.36 & 16.37 & 23.4\% & 0.8275  & 0.7203 & 13.0\% & 0.2041 & 0.2599 & 27.3\% & 0.6828 & 0.6039 & 11.6\%\\
        & 3DGM~\cite{li2024memorize} & 20.96 & 16.35 & 22.0\% & 0.8293 & \hl{\textbf{0.7248}} & 12.6\% & 0.2003 & 0.2542 &26.9\% & 0.6802 & 0.6087 & 10.5\%\\
        & GSPro~\cite{cheng2024gaussianpro} & \hl{\textbf{21.51}} & \hl{\textbf{16.39}} & 23.8\% & \hl{\textbf{0.8310}} & 0.7189 &13.5\% & \hl{\textbf{0.1804}} & \hl{\textbf{0.2450}} &35.8\% &  \hl{\textbf{0.7081}} & \hl{\textbf{0.6130}} &13.4\%\\
        & VEGS~\cite{hwang2024vegs} & 21.26 & 15.88 &25.3\% & 0.8107 & 0.7047 &13.1\% & 0.2498 & 0.3062 &22.6\% & 0.6323 & 0.5521 &12.7\% \\
        & PGSR~\cite{chen2024pgsr} & 20.57 & 16.32 & \hl{\textbf{20.7\%}} & 0.8104 & 0.7102 &12.4\% & 0.2262 & 0.2733 &20.8\% & 0.6515 & 0.5848 &10.2\%\\
        & 2DGS~\cite{huang20242d} & 20.87 & 16.30 &21.9\% & 0.8076 & 0.7103 &12.0\% & 0.2438 & 0.2890 & \hl{\textbf{18.5\%}} & 0.6256 & 0.5644 &\hl{\textbf{9.8\%}}\\
        & Feature 3DGS~\cite{zhou2024feature} & 21.02 & 16.01 & 23.8\% & 0.8096 & 0.7243 & \hl{\textbf{10.5\%}} & 0.1876 & 0.2575 & 37.3\% & 0.6958 & 0.6122 & 12.0\% \\
        & Zip-NeRF~\cite{barron2023zip} & 19.68 & 14.06 & 28.6\% & 0.7856 & 0.6917 & 12.0\% &  0.2711 & 0.3418 & 26.1\% & 0.6318 & 0.5542 & 12.3\% \\
        & Instant-NGP~\cite{mueller2022instant} & 18.77 & 12.65 & 32.6\% & 0.7631 & 0.6252 & 18.1\% & 0.4874 & 0.5938 & 21.8\% & 0.5465 & 0.4837 & 11.5\%\\
        \rowcolor{green!25} & AVERAGE & 20.67 & 15.59 & 24.6\% & 0.8083 & 0.7046 & 12.8\% & 0.2501 & 0.3134 & 25.3\% & 0.6505 & 0.5774 & 11.2\%\\

        \midrule
        \multirow{9}{*}{\textbf{Setting 2}} 
        & 3DGS~\cite{kerbl3Dgaussians} & 25.75 & 19.53 & 24.2\% & 0.8766 & 0.7511 & 14.3\% & 0.1536 & 0.2668 & 73.7\% & 0.7327 & 0.6319 & 13.8\% \\
        & 3DGM~\cite{li2024memorize} & 25.75 & 18.78 & 27.1\% & 0.8786 & 0.7464 & 15.0\% & 0.1556 & 0.2813 & 80.8\% & 0.7278 & 0.6344 & 12.8\% \\
        & GSPro~\cite{cheng2024gaussianpro} & 26.42 & 19.39 & 26.6\% & \hl{\textbf{0.8821}} & 0.7470 & 15.3\% & \hl{\textbf{0.1329}} & \hl{\textbf{0.2246}} & 69.0\% & \hl{\textbf{0.7523}} & \hl{\textbf{0.6487}} & 13.8\% \\
        & VEGS~\cite{hwang2024vegs} & 24.54 & \hl{\textbf{23.33}} & \hl{\textbf{4.9\%}} & 0.8366 & \hl{\textbf{0.7949}} & \hl{\textbf{5.0\%}} & 0.2301 & 0.2811 & \hl{\textbf{22.2\%}} & 0.6595 & 0.6133 & \hl{\textbf{7.0\%}} \\
        & PGSR~\cite{chen2024pgsr} & 24.53 & 18.38 & 25.1\% & 0.8612 & 0.7119 & 17.3\% & 0.1555 & 0.2532 & 62.8\% & 0.7200 & 0.5817 & 19.2\% \\
        & 2DGS~\cite{huang20242d} & 25.15 & 18.83 & 25.1\% & 0.8578 & 0.7204 & 16.0\% & 0.1756 & 0.2917 & 66.1\% & 0.6898 & 0.5785 & 16.1\% \\
        & Feature 3DGS~\cite{zhou2024feature} & 24.91 & 19.59 & 21.4\% & 0.8800 & 0.7864 & 10.6\% & 0.1377 & 0.2278 & 65.4\% & 0.7427 & 0.6464 & 13.0\% \\
        & Zip-NeRF~\cite{barron2023zip} & \hl{\textbf{29.06}} & 17.36 & 40.3\% & 0.8660 & 0.6715 & 22.5\% & 0.2078 & 0.3582 & 72.4\% & 0.7479 & 0.5843 & 21.9\% \\
        & Instant-NGP~\cite{mueller2022instant} & 25.61 & 17.15 & 33.0\% & 0.8596 & 0.7212 & 16.1\% & 0.3340 & 0.5171 & 54.8\% & 0.7254 & 0.6182 & 14.8\% \\
        \rowcolor{green!25} & AVERAGE & 25.75 & 19.15 & 25.6\% & 0.8665 & 0.7390 & 14.7\% & 0.1870 & 0.3002 & 62.5\% & 0.7220 & 0.6153 & 14.8\% \\

        \midrule
        \multirow{9}{*}{\textbf{Setting 3}} 
        & 3DGS~\cite{kerbl3Dgaussians} & 21.22 & \hl{\textbf{14.99}} & 29.4\% & 0.8550 & 0.7169 & 16.1\% & 0.2252 & 0.4050 & 79.8\% & 0.7002 & \hl{\textbf{0.4774}} & 31.8\% \\
        & 3DGM~\cite{li2024memorize} & 20.62 & 14.60 & 29.2\% & 0.8543 & \hl{\textbf{0.7233}} & 15.3\% & 0.2254 & 0.4049 & 79.6\% & 0.6874 & 0.4672 & 32.0\% \\
        & GSPro~\cite{cheng2024gaussianpro} & 21.58 & 14.82 & 31.3\% & 0.8634 & 0.6996 & 19.0\% & 0.2010 & 0.3877 & 92.9\% & 0.7093 & 0.4541 & 36.0\% \\
        & VEGS~\cite{hwang2024vegs} & 21.13 & 14.25 & 32.6\% & 0.8266 & 0.6475 & 21.7\% & 0.2359 & 0.4422 & 87.5\% & 0.6785 & 0.4442 & 34.5\% \\
        & PGSR~\cite{chen2024pgsr} & 19.60 & 14.17 & 27.7\% & 0.8238 & 0.6984 & 15.2\% & 0.2934 & 0.4363 & 48.7\% & 0.5867 & 0.3787 & 35.5\% \\
        & 2DGS~\cite{huang20242d} & 17.35 & 11.36 & 34.5\% & 0.7568 & 0.5447 & 28.0\% & 0.4296 & 0.5459 & \hl{\textbf{27.1\%}} & 0.3552 & 0.2327 & 34.5\% \\
        & Feature 3DGS~\cite{zhou2024feature} & \hl{\textbf{21.88}} & 14.33 & 34.5\% & \hl{\textbf{0.8643}} & 0.6386 & 26.1\% & \hl{\textbf{0.1400}} & \hl{\textbf{0.3816}} & 172.6\% & \hl{\textbf{0.7411}} & 0.4669 & 37.0\% \\
        & Zip-NeRF~\cite{barron2023zip} & 20.61 & 14.42 & 30.0\% & 0.8383 & 0.6565 & 21.7\% & 0.2197 & 0.4546 & 106.9\% & 0.7108 & 0.3645 & 48.7\% \\
        & Instant-NGP~\cite{mueller2022instant} & 19.63 & 14.39 & \hl{\textbf{26.7\%}} & 0.8179 & 0.7104 & \hl{\textbf{13.1\%}} & 0.4956 & 0.6592 & 33.0\% & 0.6083 & 0.4157 & \hl{\textbf{31.7\%}} \\
        \rowcolor{green!25} & AVERAGE & 20.40 & 14.15 & 30.6\% & 0.8334 & 0.6707 & 19.5\% & 0.2740 & 0.4575 & 70.0\% & 0.6419 & 0.4113 & 35.9\% \\
        \bottomrule
    \end{tabular}}
    \label{table:results}
    \vspace{-3mm}
\end{table*}

\section{Experiment}
\subsection{Experiment Setup}

\begin{table}[t]
\footnotesize
    \centering
    \setlength{\tabcolsep}{2.2mm}
    \caption{\textbf{Perceptual quality metric of VEGS~\cite{hwang2024vegs}.} Our benchmark also incorporates the Fréchet Inception Distance (FID) to evaluate the inpainting ability of models with generative priors.}
    \begin{tabular}{lcccc}
        \toprule
         & \textbf{Setting 1} & \textbf{Setting 2} & \textbf{Setting 3} & \textbf{AVERAGE}\\ 
        \midrule
        \textbf{FID (In.)\,$\downarrow$} & 46.3 & 33.5 & 46.1 & 42.0\\  
        \textbf{FID (Ex.)\,$\downarrow$}  & 87.4 & 67.2 & 132.4 & 95.7 \\  
        \textbf{FID (Drop)}  & 89\% & 101\% & 187\% & 126.7\%\\  
        \bottomrule
    \end{tabular}
    \label{table:FID}
    \vspace{-3mm}
\end{table}

\noindent \textbf{Implementation Details.} All 3DGS-based methods are initialized only from sparse points obtained by COLMAP~\cite{schoenberger2016sfm} and exclude lidar points. We employ Grounded-SAM-2
~\cite{ravi2024sam2segmentimages} to mask out pixels corresponding to potentially movable objects during both training and evaluation and also exclude them from the initialization of 3DGS.
\begin{figure*}[ht]
    \centering
    \includegraphics[width=0.98\textwidth]{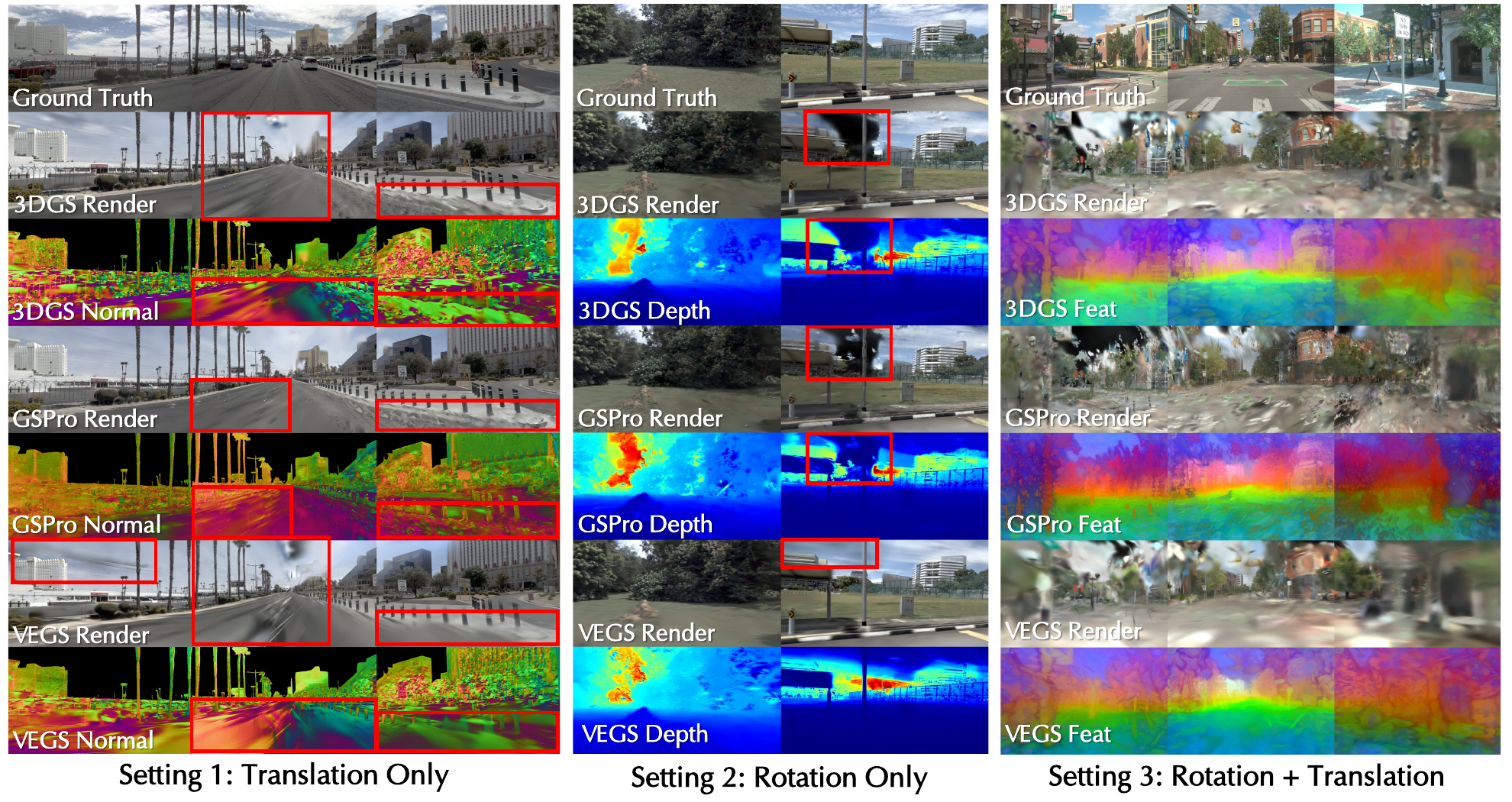}
    \caption{\textbf{Qualitative comparison of extrapolated view synthesis across different settings.} For each setting, results from different methods are compared against the ground truth. Red boxes highlight areas where methods are limited in capturing fine details, such as road surfaces, sky regions, or object boundaries, demonstrating the challenges faced by each approach under varying movement complexities.}
    \label{fig:results}

\end{figure*}

\begin{table}[t]
    \centering
    \caption{\textbf{Quantitative comparison of depth evaluation in extrapolated views in Setting 1.} 3DGM~\cite{li2024memorize} demonstrates superior performance on most evaluation metrics, while VEGS~\cite{hwang2024vegs} and GSPro~\cite{cheng2024gaussianpro} excel in SqRel and Delta1, respectively.}
    \footnotesize 
    \setlength{\tabcolsep}{4pt} 
    \setlength{\tabcolsep}{0.5mm}{
    \begin{tabular}{lcccccc}
        \toprule
        \textbf{Baseline} & \textbf{AbsRel\,$\downarrow$} & \textbf{RMSE\,$\downarrow$} & \textbf{SqRel\,$\downarrow$} & \textbf{Delta1\,$\uparrow$} & \textbf{Delta2\,$\uparrow$} & \textbf{Delta3\,$\uparrow$} \\
        \midrule
        \textbf{3DGS~\cite{kerbl3Dgaussians}} & 0.361 & 14.44 & 10.41 & 0.649 & 0.824 & 0.895 \\
        \textbf{3DGM~\cite{li2024memorize}} & \hl{\textbf{0.301}} & \hl{\textbf{13.93}} & 8.906 & 0.651 & \hl{\textbf{0.846}} & \hl{\textbf{0.915}} \\
        \textbf{PGSR~\cite{chen2024pgsr}} & 0.366 & 17.57 & 21.50 & \hl{\textbf{0.759}} & 0.834 & 0.883 \\
        \textbf{GSPro~\cite{cheng2024gaussianpro}} & 0.355 & 19.66 & 32.01 & 0.643 & 0.839 & 0.909 \\
        \textbf{VEGS~\cite{hwang2024vegs}} & 0.368 & 15.00 & \hl{\textbf{8.398}} & 0.441 & 0.691 & 0.827 \\
        \bottomrule
    \end{tabular}}
    \label{tab:depth_metrics_transposed}
    \vspace{-3mm}
\end{table}

\noindent \textbf{Evaluation Metrics.} We use three widely-used metrics to evaluate visual quality: peak signal-to-noise ratio (PSNR), structural similarity index measure (SSIM), and learned perceptual image patch similarity (LPIPS)~\cite{zhang2018unreasonable}. We also employ DINOv2~\cite{oquab2023dinov2} feature cosine similarity to evaluate image quality in latent space. For geometry evaluation, we use depth metrics, including RMSE and $\delta_{1.25}$. Additionally, we incorporate the Fréchet Inception Distance (FID)~\cite{heusel2017gans} to evaluate the impainting ability of models with strong priors.

\subsection{Experimental Results}

\autoref{table:results} presents the quantitative results across Settings 1-3, while \autoref{fig:results} illustrates the qualitative outcomes on the extrapolated test set. \autoref{table:FID} and \autoref{tab:depth_metrics_transposed} present the perceptual quality and depth metrics. The results indicate that, while the metrics perform relatively well in the interpolated test set, there is a significant drop in performance in the extrapolated test set across all baselines.

\noindent \textbf{Setting 1: Translation-only.}
In Setting 1, training views fully cover test views with moderate translational changes.
\textbf{(1)} Results show a consistent drop in performance from interpolation to extrapolation across all metrics, highlighting the challenge of generalization to views with variations. Relative drops vary by metric: PSNR falls 23–25\% (e.g., GSPro~\cite{cheng2024gaussianpro}: 21.51 $\rightarrow$ 16.39), with SSIM and LPIPS showing similar declines.
\textbf{(2)} On the extrapolative test set, methods perform comparably: GSPro tops PSNR (16.39) and minimizes LPIPS (0.2450), while 3DGM~\cite{li2024memorize} leads in SSIM (0.7248). Both yield similar feature cosine similarity; VEGS~\cite{hwang2024vegs} and PGSR~\cite{chen2024pgsr} lag slightly.
Results suggest that although GSPro marginally outperforms others, differences are small, underscoring the need for improved solutions.

\noindent \textbf{Setting 2: Rotation-only.}
In Setting 2, training views provide extensive coverage of the surrounding scene. However, most methods show poor generalization, with PSNR dropping by 22.75\%. Rotation changes are particularly challenging in textured regions (e.g., trees and intricate details), often resulting in blurring, while regions perpendicular to the vehicle remain difficult to capture. Additionally, distant areas pose reconstruction challenges, frequently leading to missing buildings and sky blackouts. Among the evaluated baselines, VEGS~\cite{hwang2024vegs} and GSPro~\cite{cheng2024gaussianpro} stand out as the best-performing baselines in this setting: VEGS utilizes its diffusion prior for inpainting and refining missing regions, while GSPro’s robust geometry handling improves generalization.

\noindent \textbf{Setting 3: Translation + Rotation.}
In Setting 3, the view changes are the largest. 
\textbf{(1)} All methods exhibit notable performance drops from interpolation to extrapolation across metrics. For instance, 3DGS~\cite{kerbl3Dgaussians} experiences a PSNR drop of 29.4\% (21.22 $\rightarrow$ 14.99), while GSPro~\cite{cheng2024gaussianpro} undergoes a similar drop of 31.3\% (21.58 $\rightarrow$ 14.82). \textbf{(2)} Feature 3DGS~\cite{zhou2024feature} and 3DGM~\cite{li2024memorize} stand out as leading methods, excelling in LPIPS (0.3816) and SSIM (0.7233), respectively. However, overall performance remains limited, with PSNR consistently falling below 15, highlighting significant room for improvement in generating high-fidelity outputs for extrapolated views.

%% file: sec/5_discussion.tex
\begin{figure*}[ht]
    \centering
    \begin{subfigure}{0.49\textwidth}
        \centering
        \includegraphics[width=\textwidth]{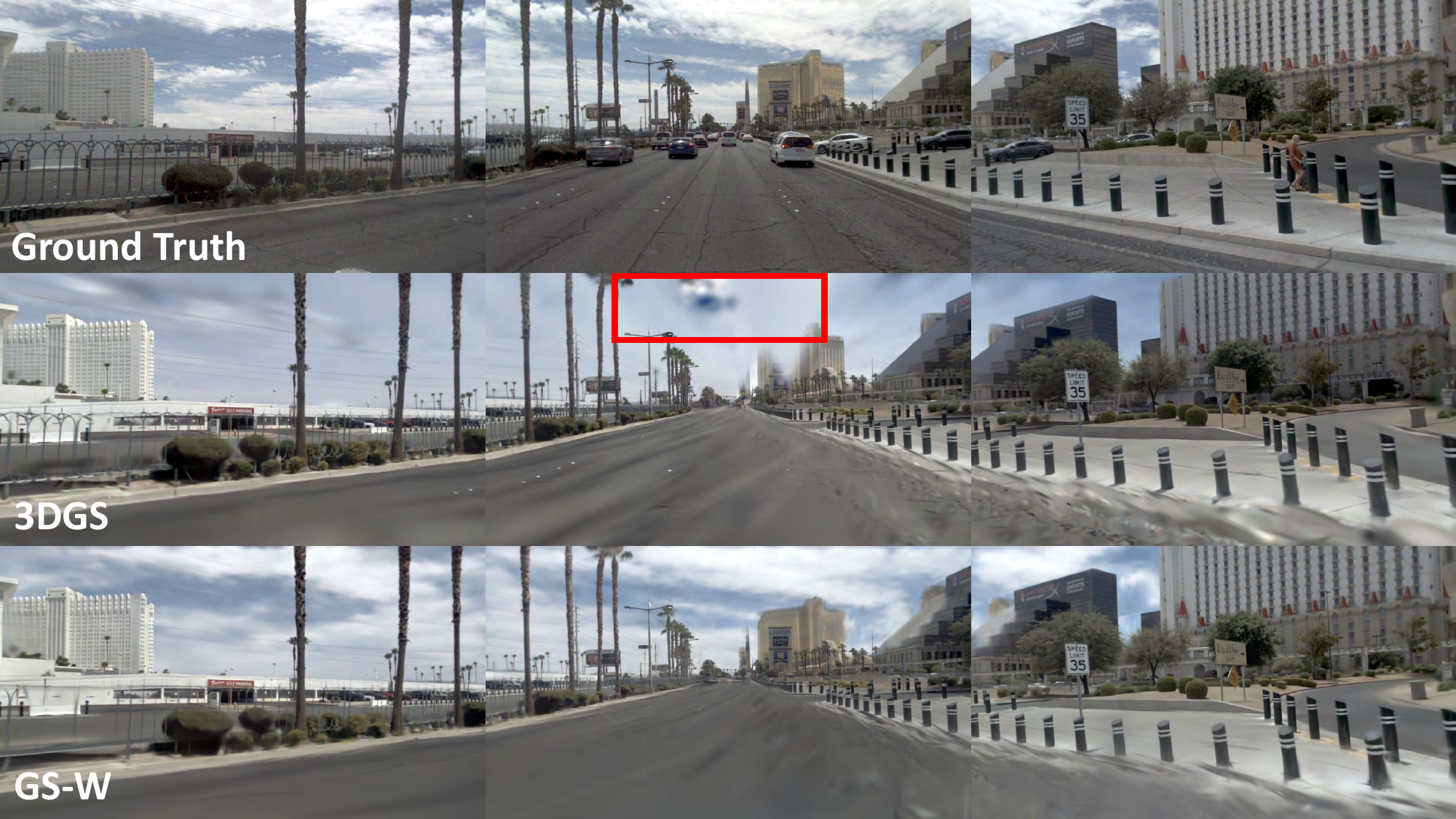}
        \caption{Mitigating lighting inconsistency by camera embedding.}
        \label{fig: gsw}
    \end{subfigure}
    \hfill
    \begin{subfigure}{0.49\textwidth}
        \centering
        \includegraphics[width=\textwidth]{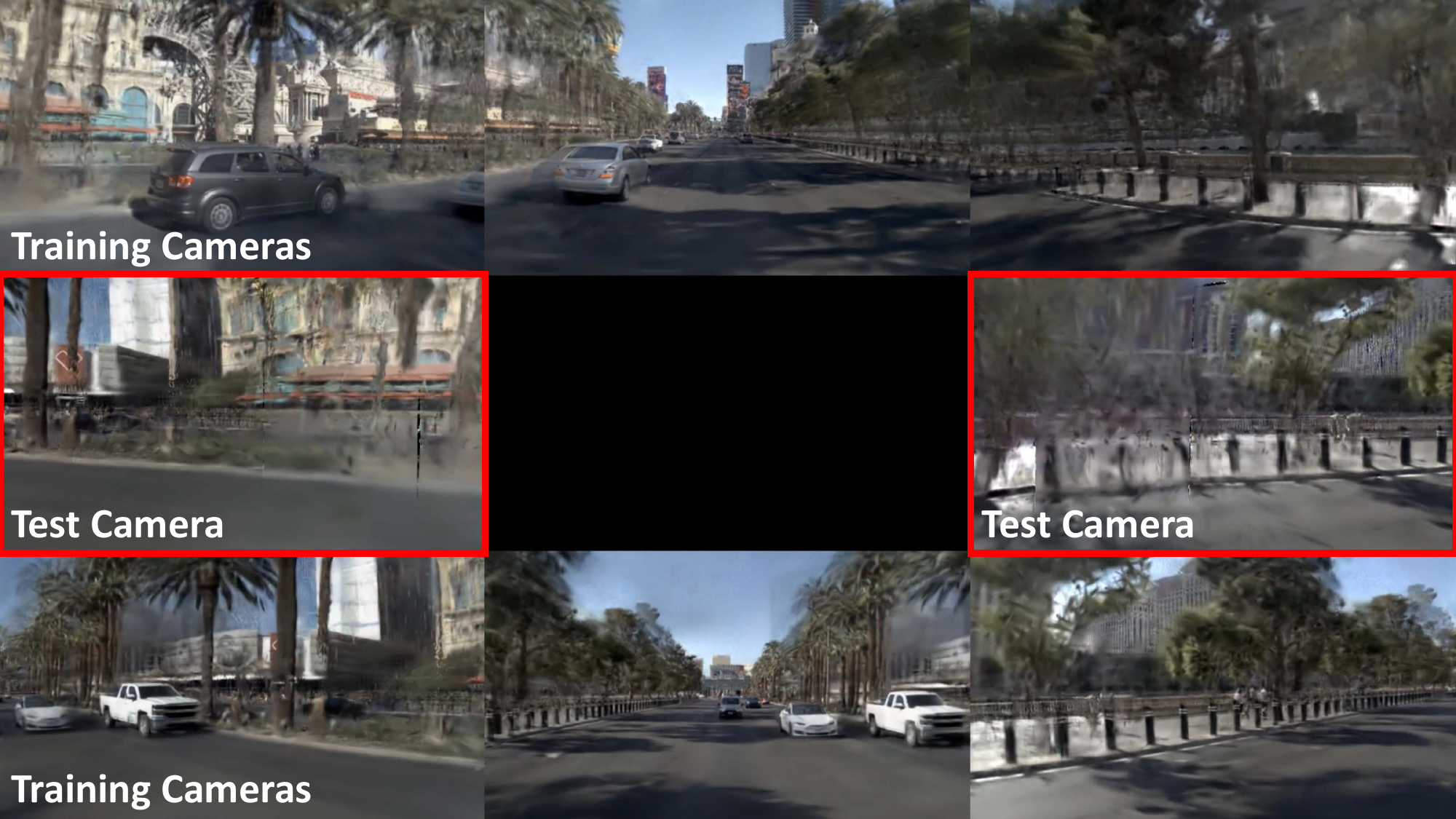}
        \caption{Qualitative dynamic scenes rendering comparison.}
        \label{fig: OmniRe}
    \end{subfigure}
    \vspace{-1.5mm}
    \caption{\textbf{Qualitative results of dynamic baseline and lighting handling.} We mitigate the lighting inconsistency issue by carefully selecting traversals under similar lighting conditions, and it can be further alleviated by introducing camera embeddings. We provide the quantitative evaluation of dynamic scene reconstruction methods in Setting 2. }
    \vspace{-3mm}
\end{figure*}

\section{Discussions} 

\noindent \textbf{Lighting Inconsistency Handling.}
Lighting inconsistencies are a widespread challenge in multi-traversal datasets due to varying illumination and weather conditions. To address this, we have taken specific measures. \textbf{(1)} Importantly, our multi-traversal dataset is manually curated to ensure that the lighting across images appears consistent to the eye. This step helps reduce the impact of extreme lighting changes and ensures a fairer evaluation of our method under controlled conditions. \textbf{(2)} One potential approach is to incorporate an appearance embedding for each image. We experiment with the Gaussian in the Wild (GS-W) baseline~\cite{zhang2024gaussian}. GS-W replaces traditional spherical harmonic-based color modeling with a method that separates intrinsic properties of each Gaussian point from dynamic appearance features of each image. This approach captures the stable, inherent appearance of objects while accommodating dynamic factors like highlights and shadows. Despite these efforts, our results, especially in Setting 3, still exhibit a significant performance drop, as shown in \autoref{table:GS-W} and \autoref{fig: gsw}, underscoring the limitations of Gaussian-based models in handling extrapolated, unseen scenarios.

\begin{table}[t]
\footnotesize
    \centering
    \caption{\textbf{Quantitative performance of GS-W~\cite{zhang2024gaussian}in different settings.} After learning the lighting features, the interpolated and extrapolated test metrics show significant improvement compared to other baselines. However, there is still a considerable drop from interpolation to extrapolation.}
    \setlength{\tabcolsep}{1.1mm}{
    \begin{tabular}{lcccccccc}
        \toprule
        \multirow{2}{*}{\textbf{Setting}} & \multicolumn{2}{c}{\textbf{PSNR}\,$\uparrow$} & \multicolumn{2}{c}{\textbf{SSIM}\,$\uparrow$} & \multicolumn{2}{c}{\textbf{LPIPS}\,$\downarrow$} & \multicolumn{2}{c}{\textbf{Feat Cos Sim}\,$\uparrow$} \\
        \cmidrule(lr){2-3} \cmidrule(lr){4-5} \cmidrule(lr){6-7} \cmidrule(lr){8-9}
         & \textbf{In.} & \textbf{Ex.} & \textbf{In.} & \textbf{Ex.} & \textbf{In.} & \textbf{Ex.} & \textbf{In.} & \textbf{Ex.} \\
        \midrule
        Setting 1 & \hl{\textbf{28.15}} & 20.22 & \hl{\textbf{0.89}} & 0.78 & \hl{\textbf{0.15}} & 0.23 & \hl{\textbf{0.71}} & 0.64 \\
        Setting 2 & \hl{\textbf{30.10}} & 21.21 & \hl{\textbf{0.91}} & 0.82 & \hl{\textbf{0.13}} & 0.20 & \hl{\textbf{0.76}} & 0.67 \\
        Setting 3 & \hl{\textbf{28.62}} & 19.36 & \hl{\textbf{0.87}} & 0.73 & \hl{\textbf{0.15}} & 0.31 & \hl{\textbf{0.74}} & 0.50 \\
        \bottomrule
    \end{tabular}}
    \label{table:GS-W}
    \vspace{-3mm}
\end{table}

\noindent \textbf{Dynamic Scenes.} In addition to the static scene reconstruction methods that we evaluated in \autoref{table:results}, some dynamic scene reconstruction techniques have recently emerged. We conduct evaluation for Setting 2 to assess these approaches, e.g., OmniRe~\cite{chen2024omnire}. It organizes rigid-deformable nodes and background nodes to capture dynamic scene structures and employs SMPL~\cite{SMPL:2015} for non-rigid object modeling. As illustrated in \autoref{fig: OmniRe}, the rendering results reveal a significant performance gap between the training and extrapolated test cameras. In the extrapolated test views, objects such as trees and stakes lose texture and geometric details, resulting in noticeably blurry outputs, whereas the training views maintain high fidelity. As shown in \autoref{table:OmniRe}, the reconstruction metrics indicate an average drop of 25\% when transitioning from interpolated to extrapolated settings. The results highlight the challenges of extrapolated view synthesis in dynamic scenes and the need for further research.

\begin{table}[t]
\footnotesize
    \centering
    \caption{\textbf{Quantitative performance of OmniRe~\cite{chen2024omnire} in Setting 2.} The experiment uses a single traversal with dynamic objects, showing a noticeable drop from interpolation to extrapolation.}
    \setlength{\tabcolsep}{1.1mm}{
    \begin{tabular}{lcccccccc}
        \toprule
        \multirow{2}{*}{\textbf{Setting}} & \multicolumn{2}{c}{\textbf{PSNR}\,$\uparrow$} & \multicolumn{2}{c}{\textbf{SSIM}\,$\uparrow$} & \multicolumn{2}{c}{\textbf{LPIPS}\,$\downarrow$} & \multicolumn{2}{c}{\textbf{Feat Cos Sim}\,$\uparrow$} \\
        \cmidrule(lr){2-3} \cmidrule(lr){4-5} \cmidrule(lr){6-7} \cmidrule(lr){8-9}
         & \textbf{In.} & \textbf{Ex.} & \textbf{In.} & \textbf{Ex.} & \textbf{In.} & \textbf{Ex.} & \textbf{In.} & \textbf{Ex.} \\
        \midrule
        Setting 2 &  \hl{\textbf{19.78}} & 15.32 & \hl{\textbf{0.65}} & 0.45 &  \hl{\textbf{0.38}} & 0.53 & \hl{\textbf{0.73}} & 0.58  \\
        \bottomrule
    \end{tabular}}
    \label{table:OmniRe}
    \vspace{-3mm}
\end{table}

%% file: sec/6_conclusion.tex
\section{Conclusions, Limitations and Future Work}
\noindent \textbf{Conclusions.} We introduce the first benchmark enabling quantitative evaluation of extrapolated view synthesis, advancing photorealistic simulation for self-driving and robotics. Our benchmark integrates real-world multi-traversal, multi-agent, and multi-camera data, categorizes scenes into different evaluation settings, and evaluates state-of-the-art NVS models. Experimental results reveal that while some methods address specific challenges, current models demonstrate limited generalization, with significant overfitting to training views and suboptimal performance in extrapolated view synthesis. To support further research, we will release the dataset and benchmark, addressing the long-standing data scarcity and providing evaluation protocols. We believe the EUVS benchmark will catalyze meaningful advancements in self-driving and robotics innovation.

 \noindent \textbf{Limitations and Future Work.} Our benchmark does have limitations. First, although we provide the ground truth to enable quantitative evaluation for static scenes in all settings and the foreground objects evaluation in Setting 2, we lack the foreground evaluation in Settings 1 and 3. Future work will aim to expand our work to include the dynamic object evaluation in Settings 1 and 3 using multi-agent data. Secondly, although we carefully manually selected and ensured that the test trajectory viewpoints are well covered by training trajectories, a more in-depth evaluation of how observed and unseen regions differ is left for future work.

\noindent\textbf{Acknowledgment.} This work was supported in part through NSF grants 2238968 and 2121391, and the NYU IT High Performance Computing resources, services, and staff expertise. Yiming Li is supported by NVIDIA Graduate Fellowship (2024-2025).

%% file: sec/X_supplementary.tex
\clearpage
\setcounter{page}{1}
\maketitlesupplementary
\renewcommand{\thetable}{\Roman{table}}
\renewcommand{\thefigure}{\Roman{figure}}
\renewcommand\thesection{\Alph {section}}
\setcounter{section}{0}
\setcounter{figure}{0}
\setcounter{table}{0}

\begin{figure*}[ht]
    \centering
    \begin{subfigure}{0.49\textwidth}
        \centering
        \includegraphics[width=\textwidth]{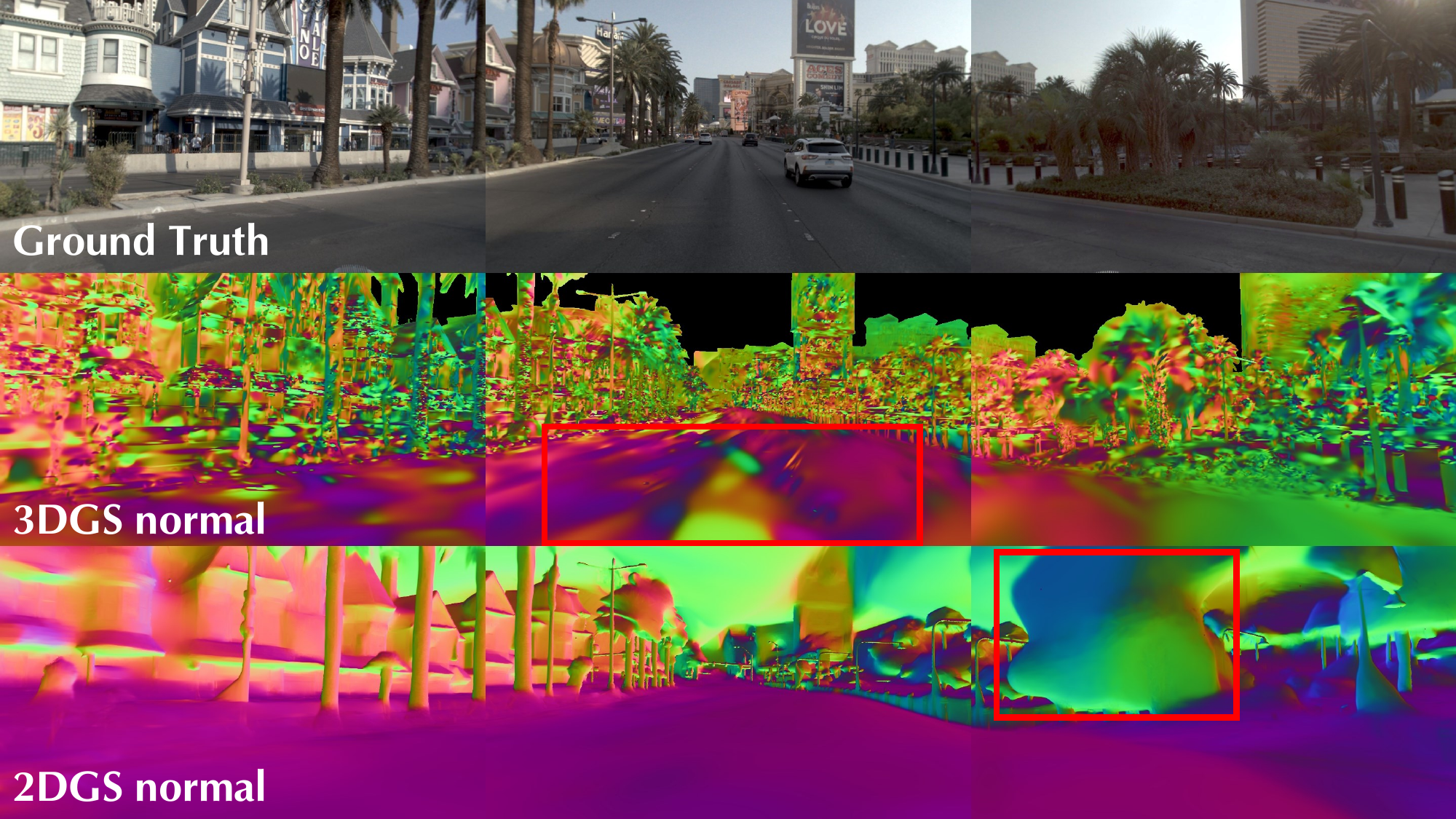}
        \caption{Planar-based vs. ellipsoid-base method.}
        \label{fig: Planar-based vs Ellipsoid-base method}
    \end{subfigure}
    \hfill
    \begin{subfigure}{0.49\textwidth}
        \centering
        \includegraphics[width=\textwidth]{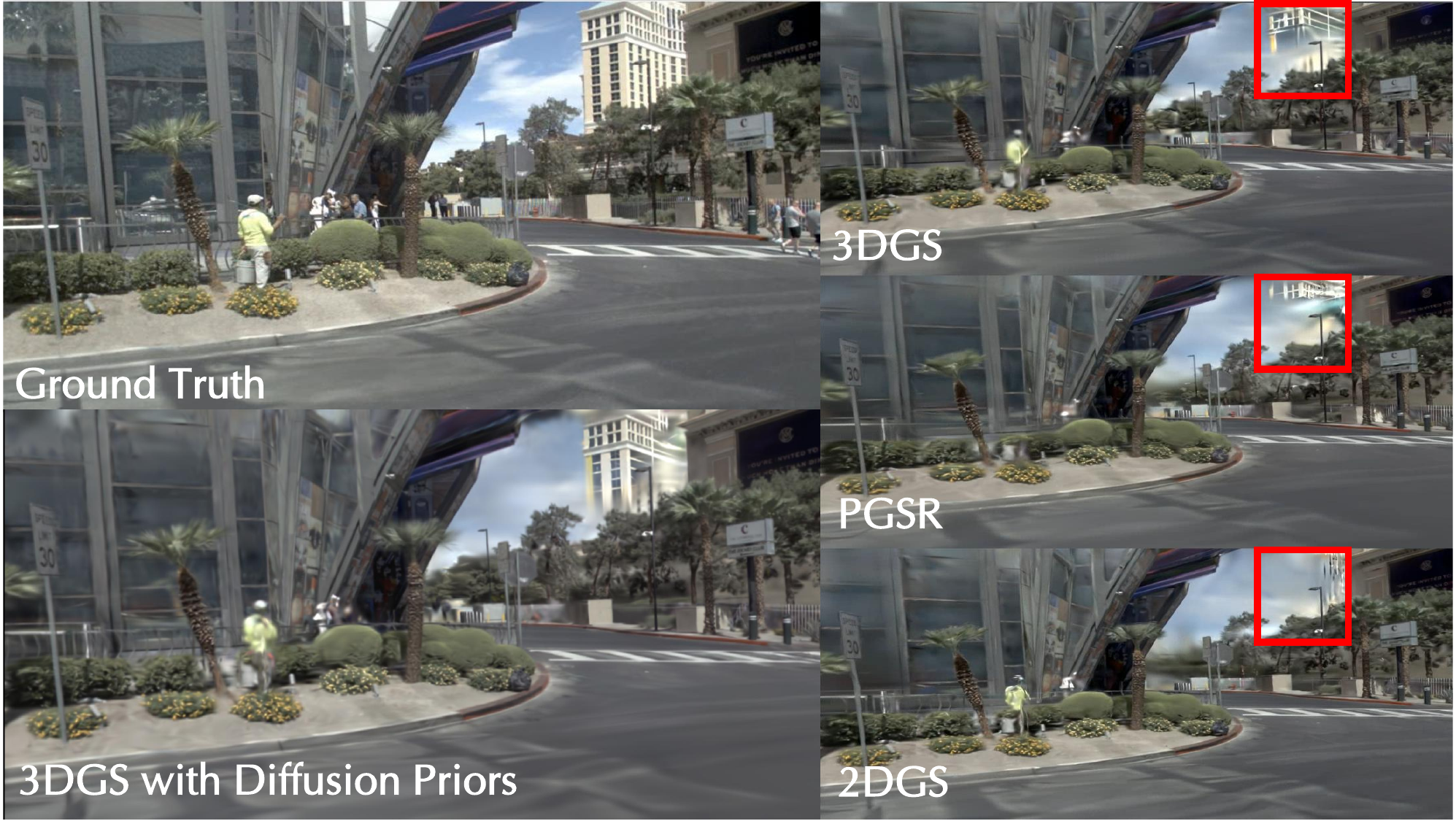}
        \caption{With vs. without diffusion priors.}
        \label{fig: Diffusion figure}
    \end{subfigure}
    \begin{subfigure}{0.49\textwidth}
        \centering
        \includegraphics[width=\textwidth]{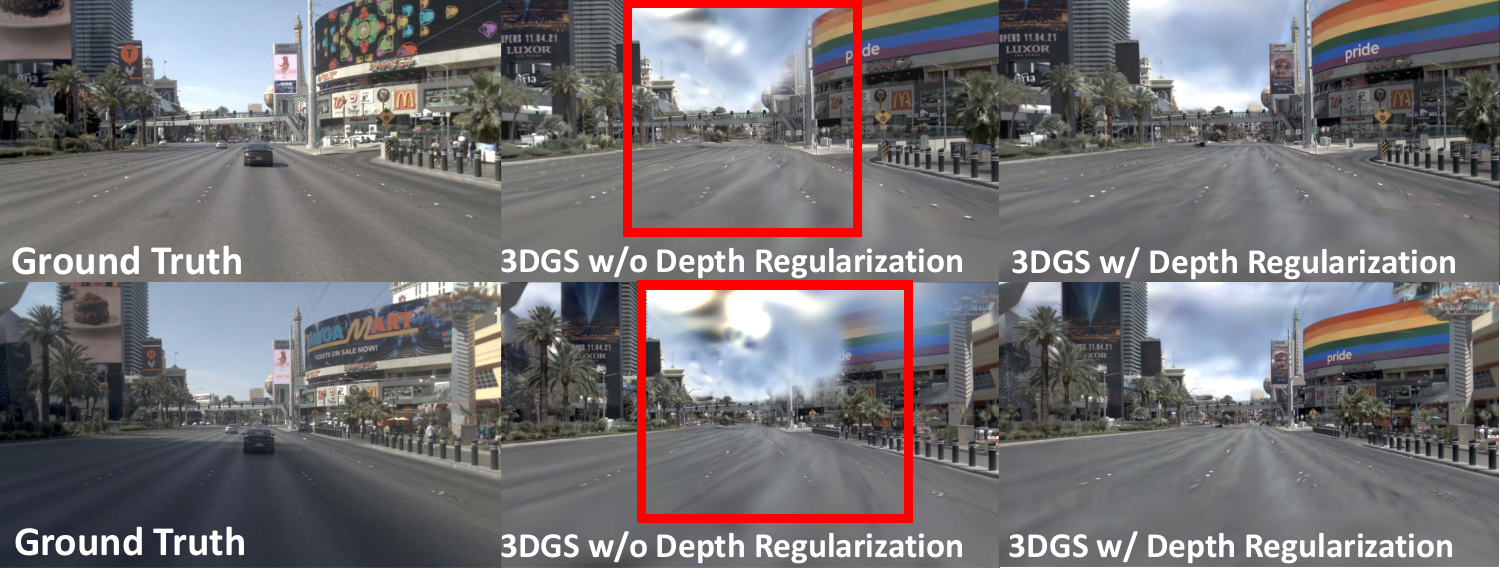}
        \caption{With vs. without depth priors.}
        \label{fig: Depth priors}
    \end{subfigure}
    \hfill
    \begin{subfigure}{0.49\textwidth}
        \centering
        \includegraphics[width=\textwidth]{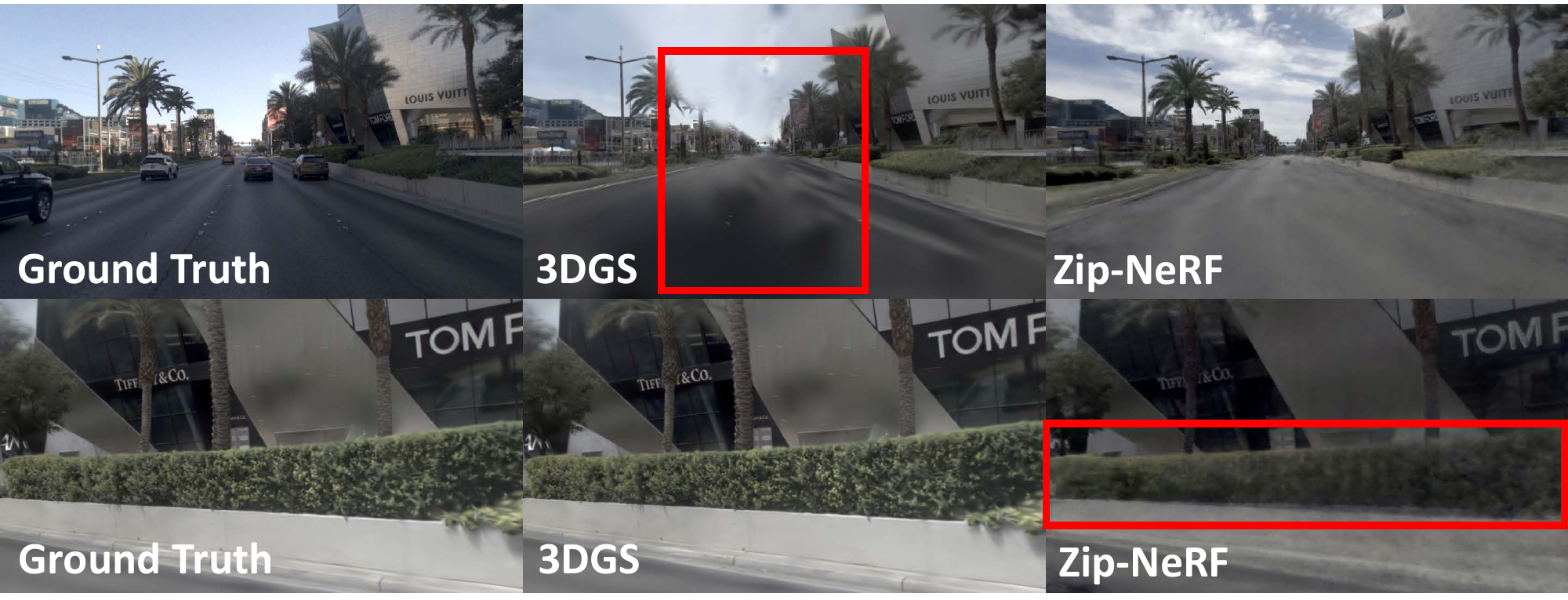}
        \caption{GS-based vs. NeRF-based.}
        \label{fig: Nerf vs GS figure}
    \end{subfigure}
    \vspace{-2mm}
    \caption{\textbf{Qualitative comparison of different techniques.} The various techniques excel in different aspects, showing some trade-offs in extrapolated view synthesis. Although they can partially address the challenges, they fail to resolve the underlying issues fundamentally.}
\end{figure*}

\section*{Appendix A: Methods Discussions}
\noindent\textbf{Planar-Based vs. Ellipsoid-Based.} Planar-based methods (e.g., GSPro~\cite{cheng2024gaussianpro}, PGSR~\cite{chen2024pgsr}, and 2DGS~\cite{huang20242d}) excel in road representation due to their planar geometry and refinement strategies but struggle with fine-textured urban objects like plants and fences. Conversely, ellipsoid-based methods (e.g., 3DGS~\cite{kerbl3Dgaussians} and 3DGM~\cite{li2024memorize}) better handle high-textured objects but often overfit, leading to errors in road representation. For instance, in the translation setting (\autoref{fig: Planar-based vs Ellipsoid-base method}), planar-based methods struggle with plants, while ellipsoid-based methods perform poorly on roads. A hybrid representation could effectively combine the strengths of both approaches to address these challenges in EUVS.

\noindent\textbf{Enhancing View Synthesis with Diffusion Priors.} While training cameras may collectively cover the entire scene, the limited number of viewpoints often results in insufficient representation of certain areas. Leveraging diffusion priors proves to be an effective approach in such cases. By supervising augmented views with diffusion priors, unseen or poorly represented views can be generated and corrected. For instance, as shown in \autoref{fig: Diffusion figure}, the building rendered by other models is fragmented, but guiding with diffusion priors helps complete the building structure and presents a holistic urban scene. 
On average, in Table 1 of the main paper, VEGS~\cite{hwang2024vegs} with diffusion priors significantly outperforms 3DGS~\cite{kerbl3Dgaussians} in the rotation-only setting, achieving a 19.4\% increase in PSNR (23.33 vs. 19.53) and a 5.8\% improvement in SSIM (0.7949 vs. 0.7511).

\noindent\textbf{Regularization by Depth Priors.} Utilizing depth priors from foundation models, such as Depth Anything~\cite{yang2024depth}, has proven to be an effective approach for enhancing training regularization~\cite{chung2024depth}. In our experiments, depth regularization enhances geometric accuracy by utilizing depth information to constrain Gaussians in regions like the sky and road to more geometrically consistent positions. As shown in \autoref{fig: Depth priors}, the sky is accurately constrained to a distant position, ensuring it does not overlap with the building during lane changes. Similarly, the road is aligned to a consistent plane, effectively mitigating the distortion issues observed in the vanilla baseline. The regularization by depth priors ensures spatial consistency and reduces visual artifacts, leading to more reasonable extrapolated views.

\noindent\textbf{Gaussian-Based vs. NeRF-Based Methods.} A fundamental difference between Gaussian-based and NeRF-based approaches lies in their representation: Gaussian-based methods rely on explicit representations, whereas NeRF-based methods use implicit representations. Our experiments reveal that implicit methods, such as Zip-NeRF~\cite{barron2023zip}, preserve overall geometry more consistently under large shifts, though it can still lose some sharpness even with small viewpoint extrapolations. In contrast, the explicit representation of Gaussian Splatting-based methods excels in regions with accurate geometry, producing sharper fine details (e.g., foliage), but struggles with incomplete geometry under large shifts. , as illustrated in~\autoref{fig: Nerf vs GS figure}.

\begin{figure}[ht]
    \centering
    \includegraphics[width=\linewidth]{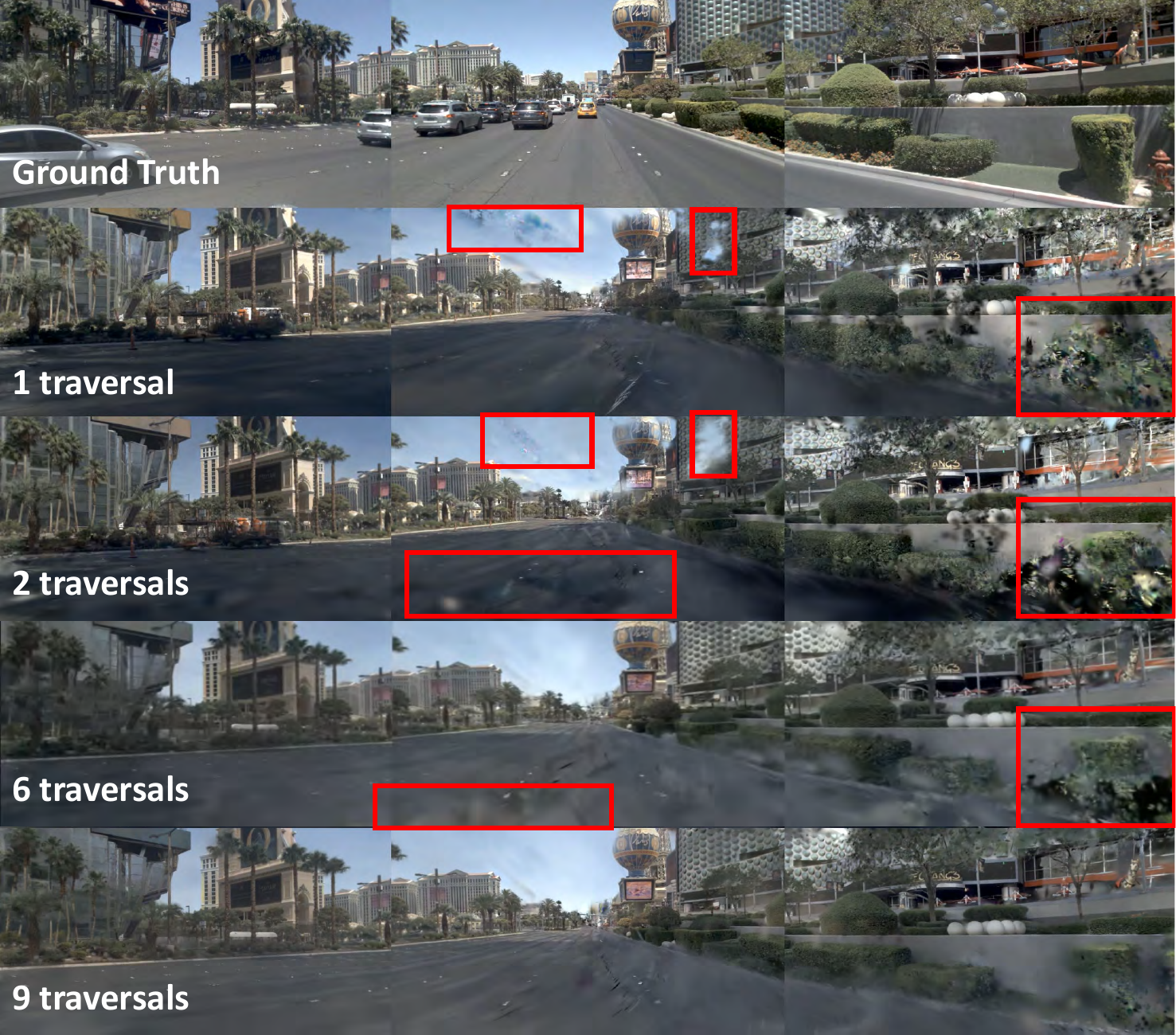}
    \caption{\textbf{As the number of traversals increases, the performance of NVS improves.} This is highlighted in the red box, where the texture progressively enriches and errors in areas like the sky and ground are reduced.}
    \label{fig:number of traversals}
    \vspace{-3mm}
\end{figure}

\begin{figure}[ht]
    \centering
    \includegraphics[width=\linewidth]{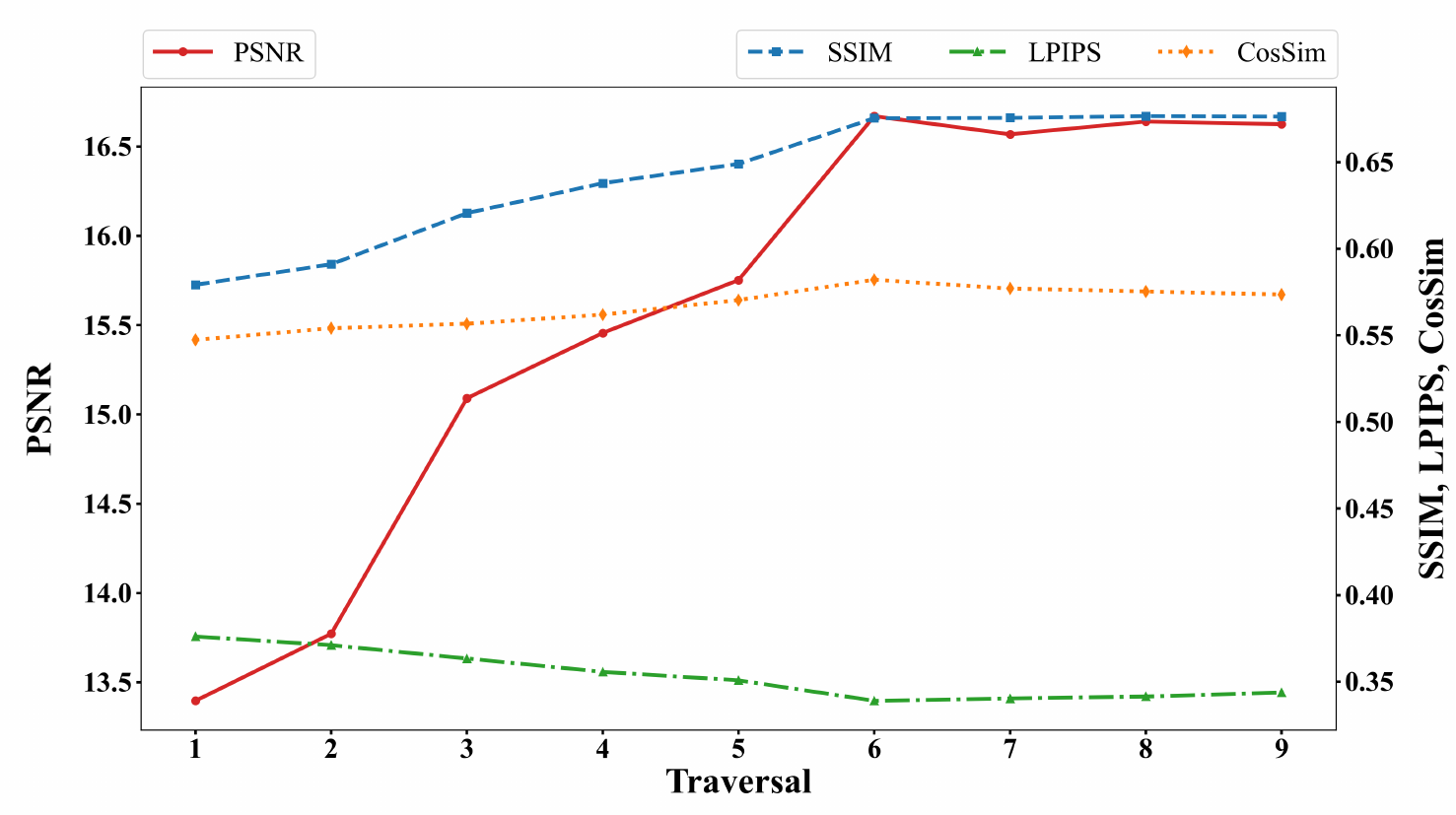}
    \caption{\textbf{NVS performance vs. number of traversals.} With more traversals, PSNR and SSIM exhibit notable improvements, indicating enhanced image quality and structural similarity. LPIPS values decrease, reflecting better perceptual consistency, while CosSim stabilizes after an initial rise. These results highlight the importance of more visual data for improving NVS performance.}
    \label{fig:Metric performance vs. number of traversals}
    \vspace{-3mm}
\end{figure}

\noindent \textbf{Performance Gains from Multi-Traversal Data.} Multi-traversal data plays a critical role in Extrapolated View Synthesis. Using the GaussianPro model~\cite{cheng2024gaussianpro} in Setting 1, we progressively increase the number of training traversals to observe its impact. The results, shown in \autoref{fig:Metric performance vs. number of traversals} and \autoref{fig:number of traversals}, indicate that as the number of traversals increases, the NVS metrics for the test view gradually improve, then plateau. This consistent improvement stems from increased unique observations, enabling diverse perspectives and more accurate background reconstruction while reducing dynamic object influence. This suggests that incorporating more visual data can help improve the performance of extrapolated view synthesis.

\section*{Appendix B: Comparison of Baselines}

\begin{figure*}[!b]
    \centering
    \begin{subfigure}{0.74\textwidth}
        \centering
        \includegraphics[width=\textwidth]{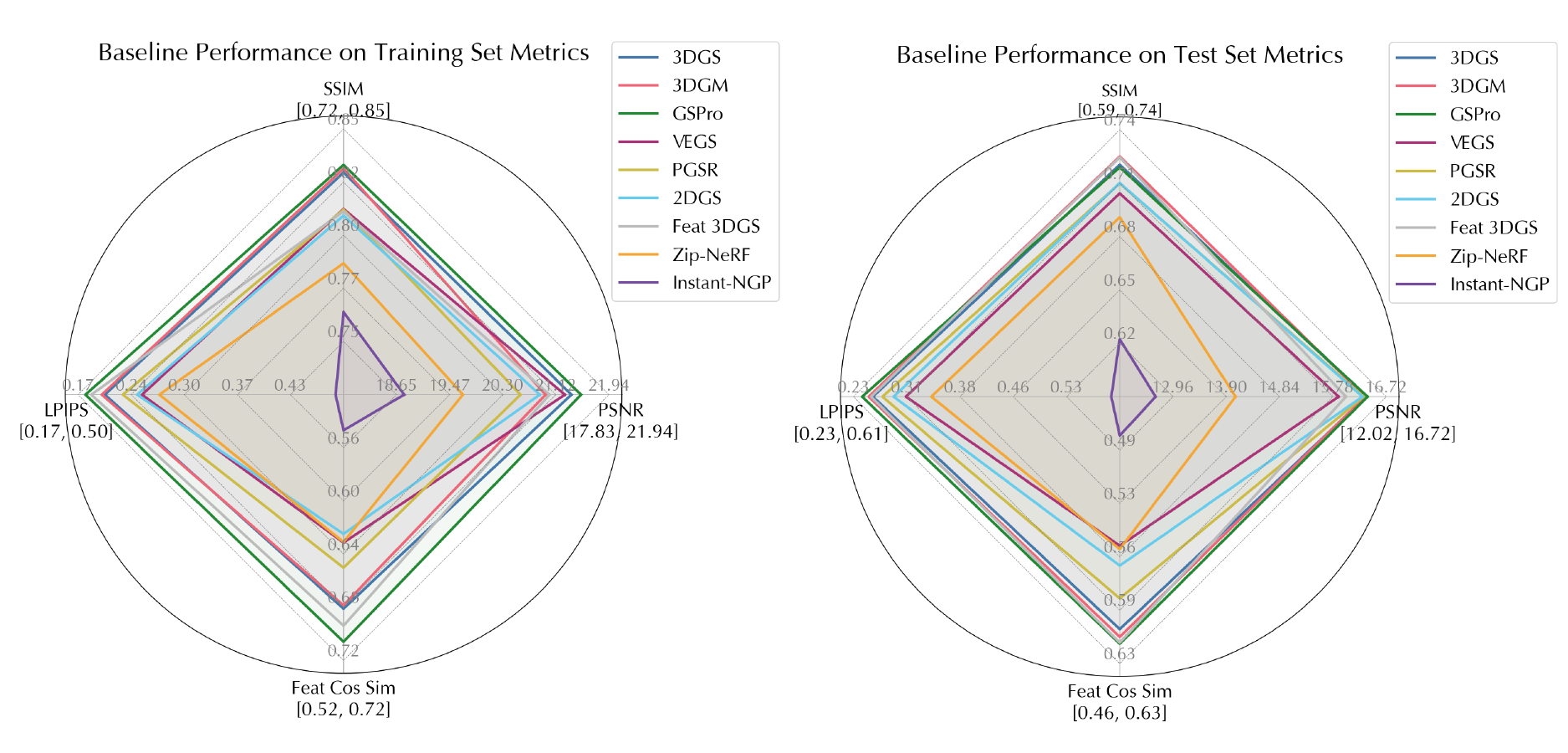}
        \caption{Baseline performance comparison in Setting 1.}
        \label{fig: Setting1_baseline_radar}
    \end{subfigure}
    \hfill
    \begin{subfigure}{0.74\textwidth}
        \centering
        \includegraphics[width=\textwidth]{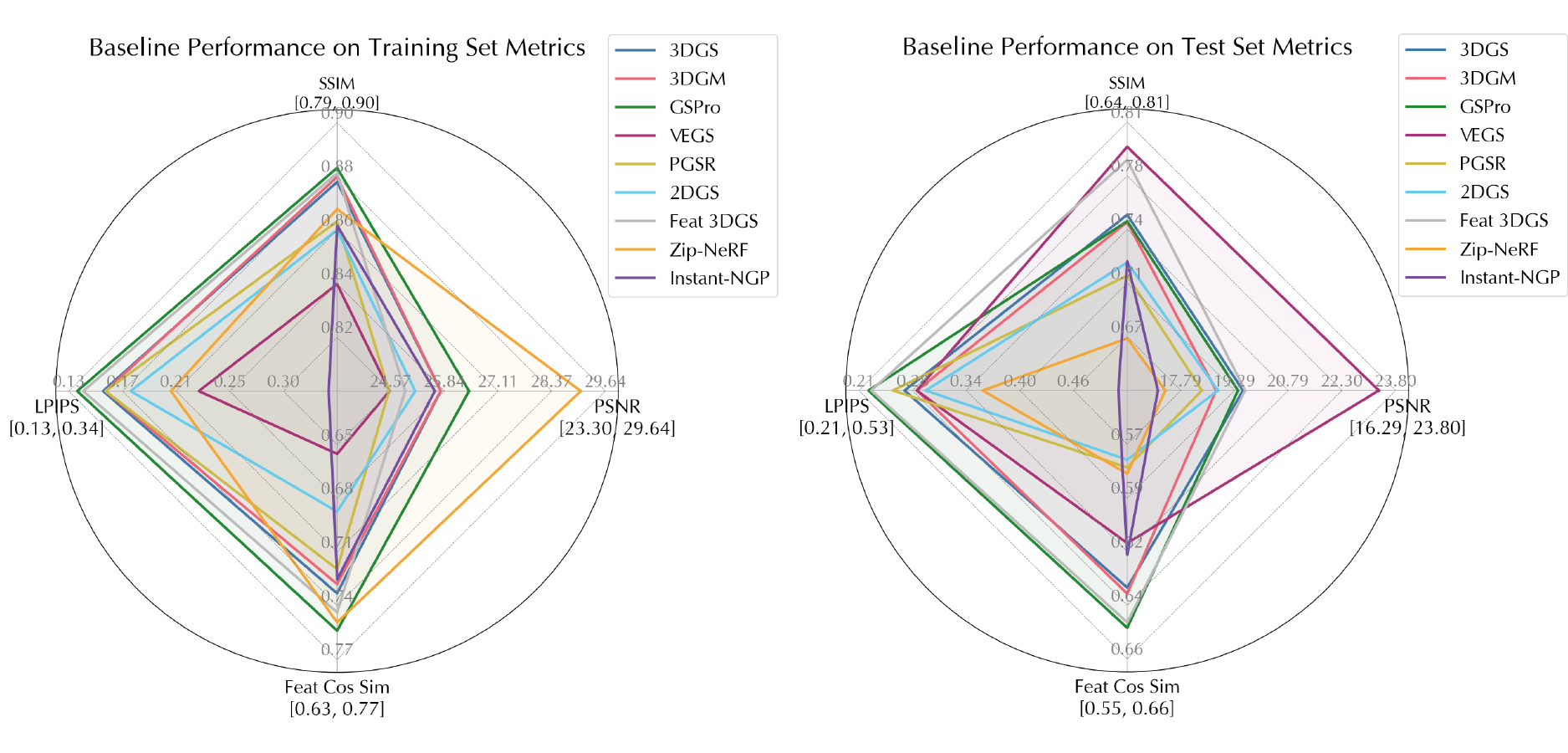}
        \caption{Baseline performance comparison in Setting 2.}
        \label{fig: Setting2_baseline_radar}
    \end{subfigure}
    \begin{subfigure}{0.74\textwidth}
        \centering
        \includegraphics[width=\textwidth]{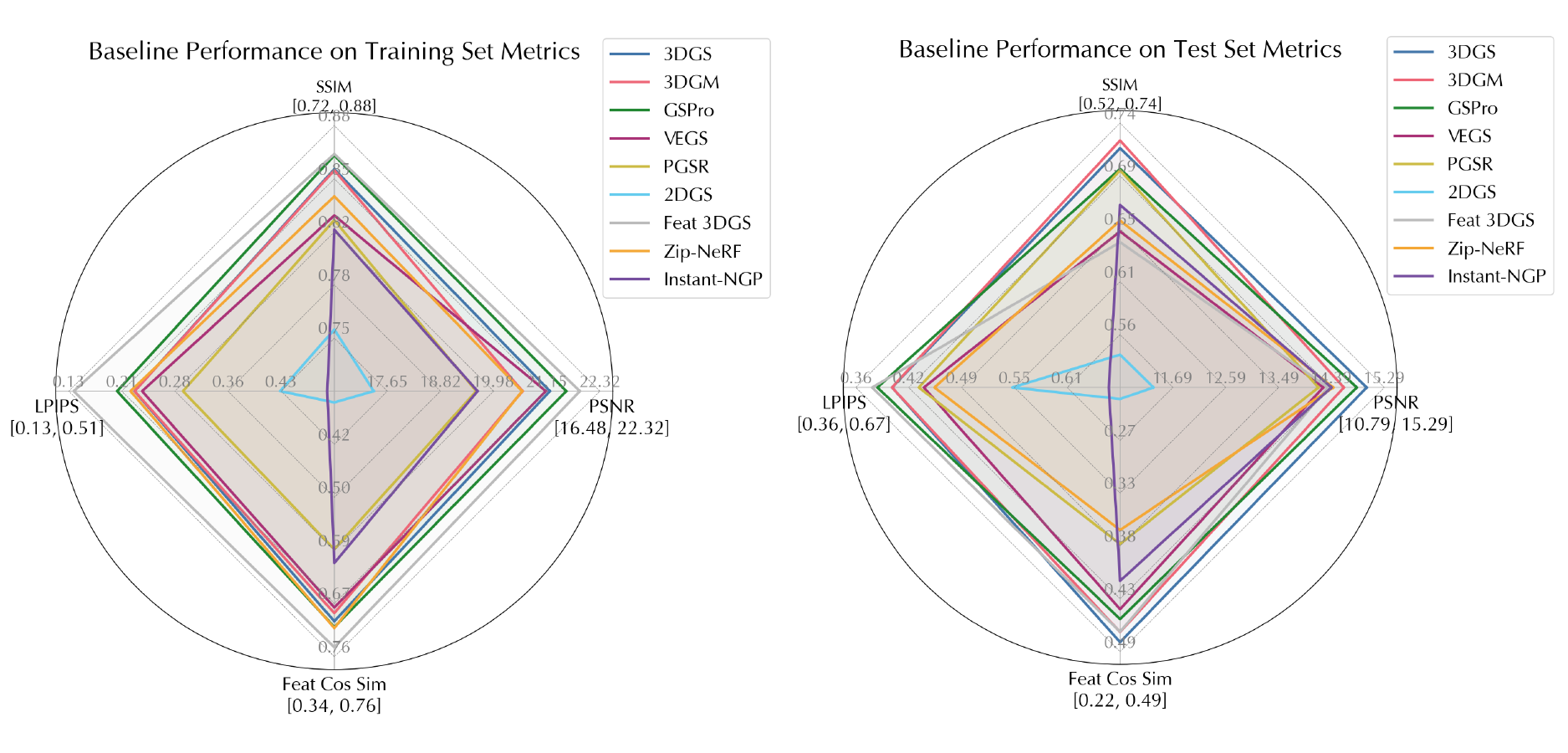}
        \caption{Baseline performance comparison in Setting 3.}
        \label{fig: Setting3_baseline_radar}
    \end{subfigure}
    \caption{\textbf{Baseline performance comparison across different settings.} Since scenes in different settings evaluate varying capabilities, different baselines demonstrate strengths in different evaluation settings.}
    \label{fig:Baseline performance comparison across different settings}
\end{figure*}

\noindent \textbf{Quantitative Comparison.} We report the quantitative performance comparison across all settings and baselines in \autoref{fig:Baseline performance comparison across different settings}. \textbf{(1)} In Setting 1, the performance gaps on the extrapolative test set are small, with most baselines performing comparably poorly. Among them, 3DGS~\cite{kerbl3Dgaussians}, 3DGM~\cite{li2024memorize}, and GSPro achieve relatively better results. \textbf{(2)} In Setting 2 (\autoref{fig: Setting2_baseline_radar}), in extrapolated views, VEGS~\cite{hwang2024vegs} significantly outperforms all other methods, achieving at least 20\% higher PSNR. These results highlight the effectiveness of diffusion priors in rotation-only settings. \textbf{(3)} In Setting 3, as shown in \autoref{fig: Setting3_baseline_radar}, none of the baselines exhibit a clear advantage, as all methods fail equally in this challenging setting. On the extrapolative test set, different baselines exhibit strengths in specific metrics, but no method demonstrates superiority across all metrics, indicating that all baselines struggle with extrapolated view synthesis and fail to address it fundamentally.

\begin{figure*}[ht]
    \centering
    \begin{subfigure}{0.49\textwidth}
        \centering        
        \includegraphics[width=\textwidth]{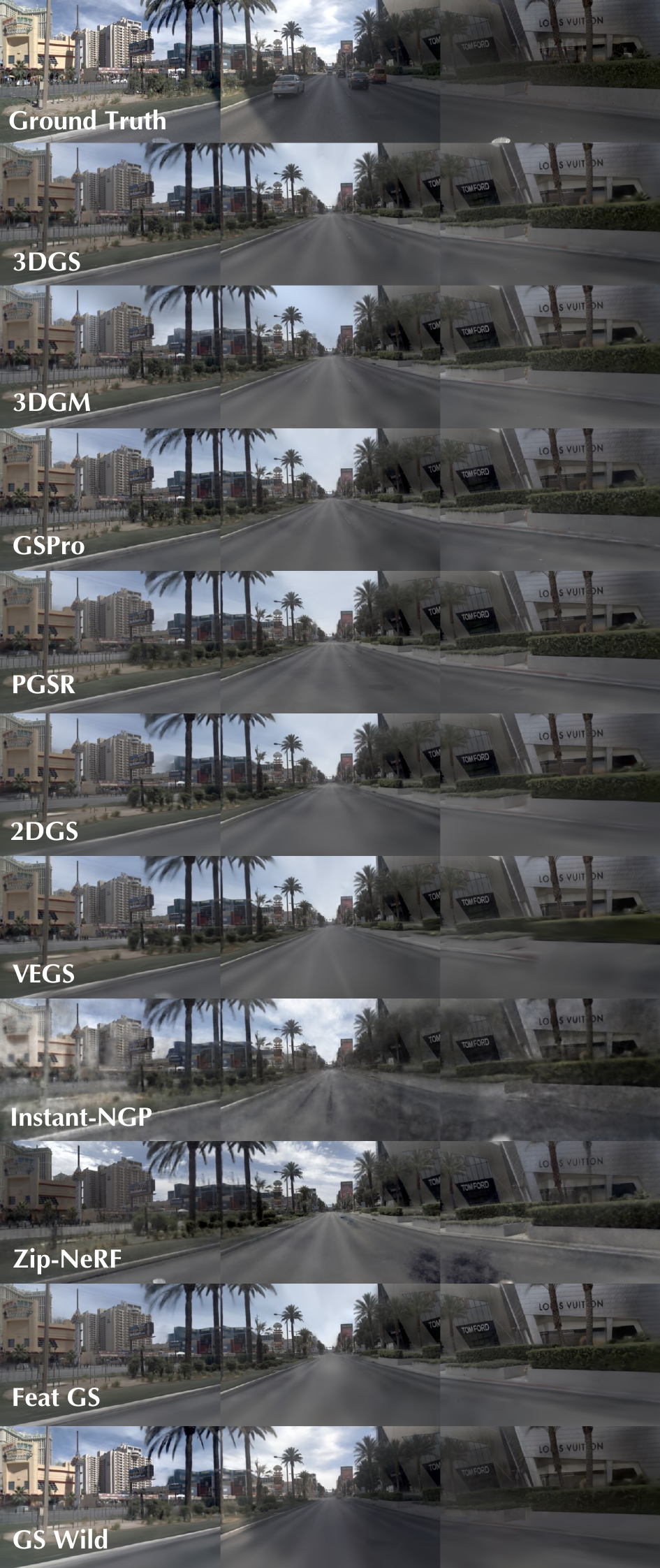}
        \caption{Rendering results comparison in original view.}
        \label{fig:Detailed results setting 1 train}
    \end{subfigure}
    \hfill
    \begin{subfigure}{0.49\textwidth}
        \centering \includegraphics[width=\textwidth]{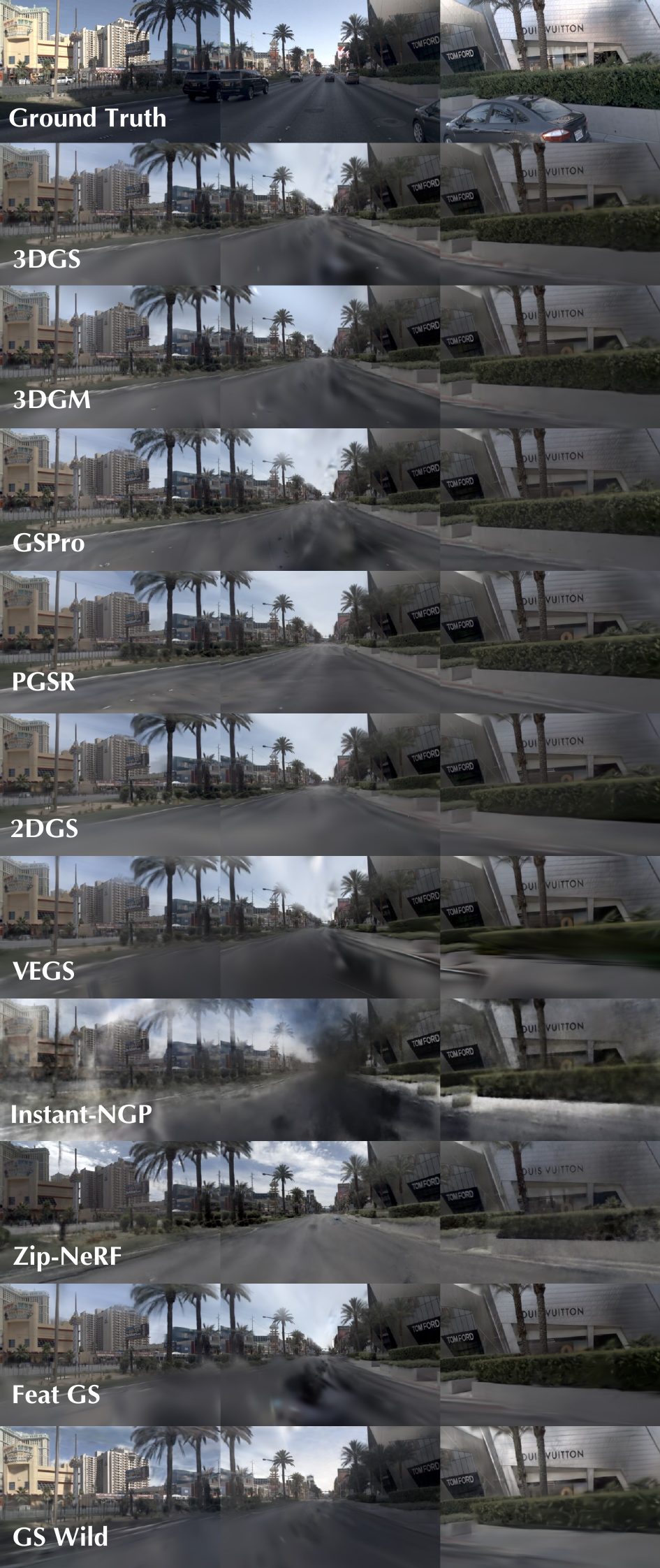}
        \caption{Rendering results comparison in extrapolated view.}
        \label{fig: Detailed results setting 1 test}
    \end{subfigure}
    \caption{\textbf{Qualitative comparison of baseline methods in Setting 1.} Ground reconstruction failures and floating artifacts in the sky are particularly noticeable, highlighting the challenges in the lane change.}
    \label{fig:Qualitative results comparison across all baselines at setting 1}
\end{figure*}

\begin{figure*}[ht] 
    \centering
    \includegraphics[width=0.95\textwidth]{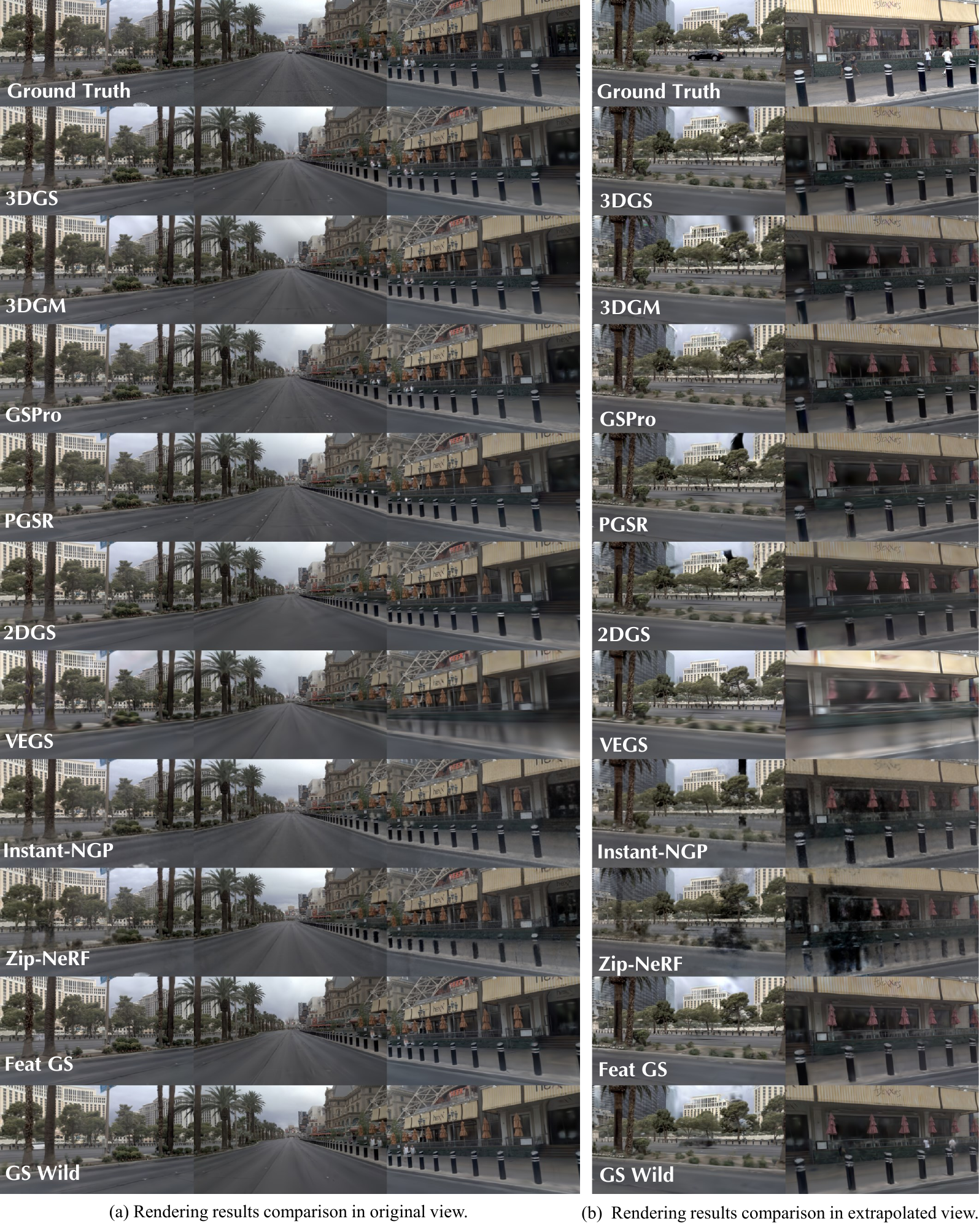}
    \caption{\textbf{Qualitative comparison of baseline methods in Setting 2.} The three front and three back cameras (six in total) are used for training, while the two side cameras are reserved for testing. To ensure clarity and conciseness, only a subset of the training cameras is visualized here due to space limitations.}
    \label{fig:Qualitative results comparison across all baselines at setting 2}
\end{figure*}

\begin{figure*}[ht]
    \centering
    \begin{subfigure}{0.49\textwidth}
        \centering        
        \includegraphics[width=\textwidth]{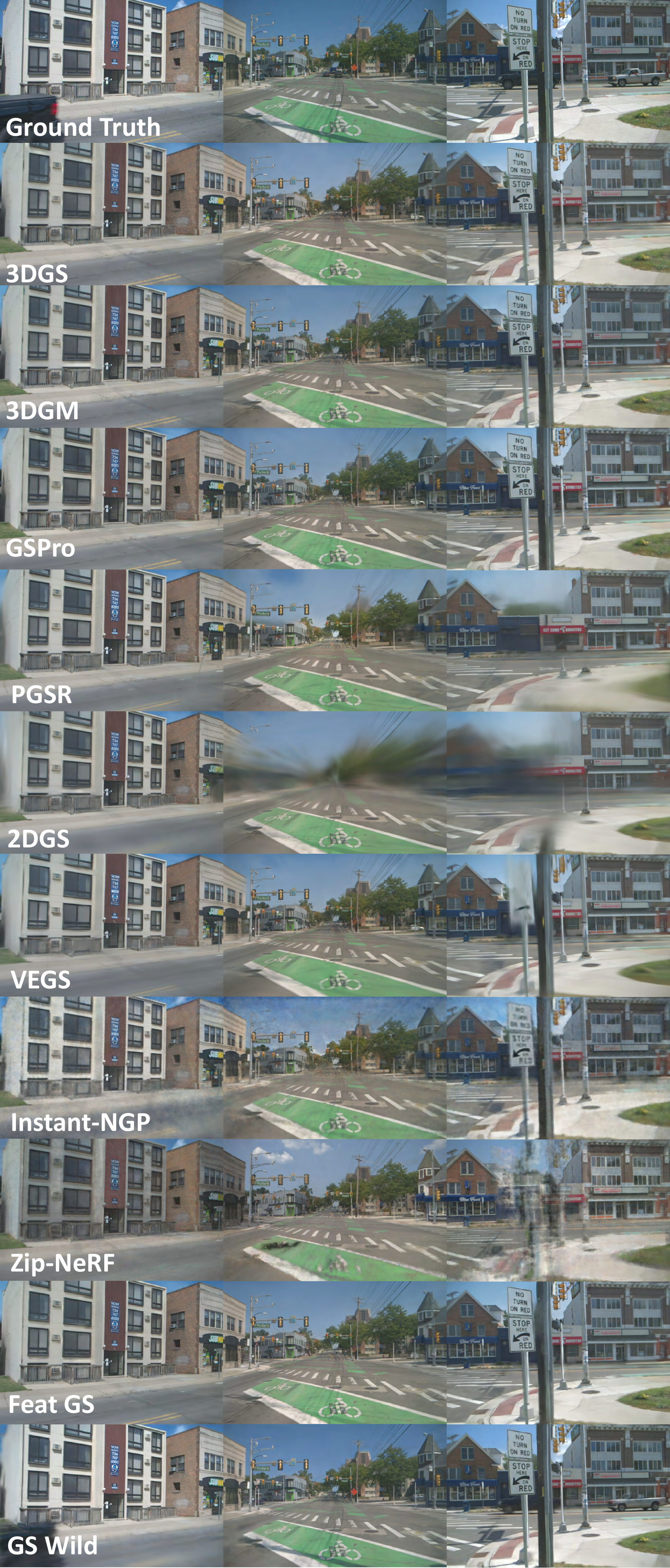}
        \caption{Rendering results comparison in original view.}
        \label{fig:Detailed results setting 3 train}
    \end{subfigure}
    \hfill
    \begin{subfigure}{0.49\textwidth}
        \centering \includegraphics[width=\textwidth]{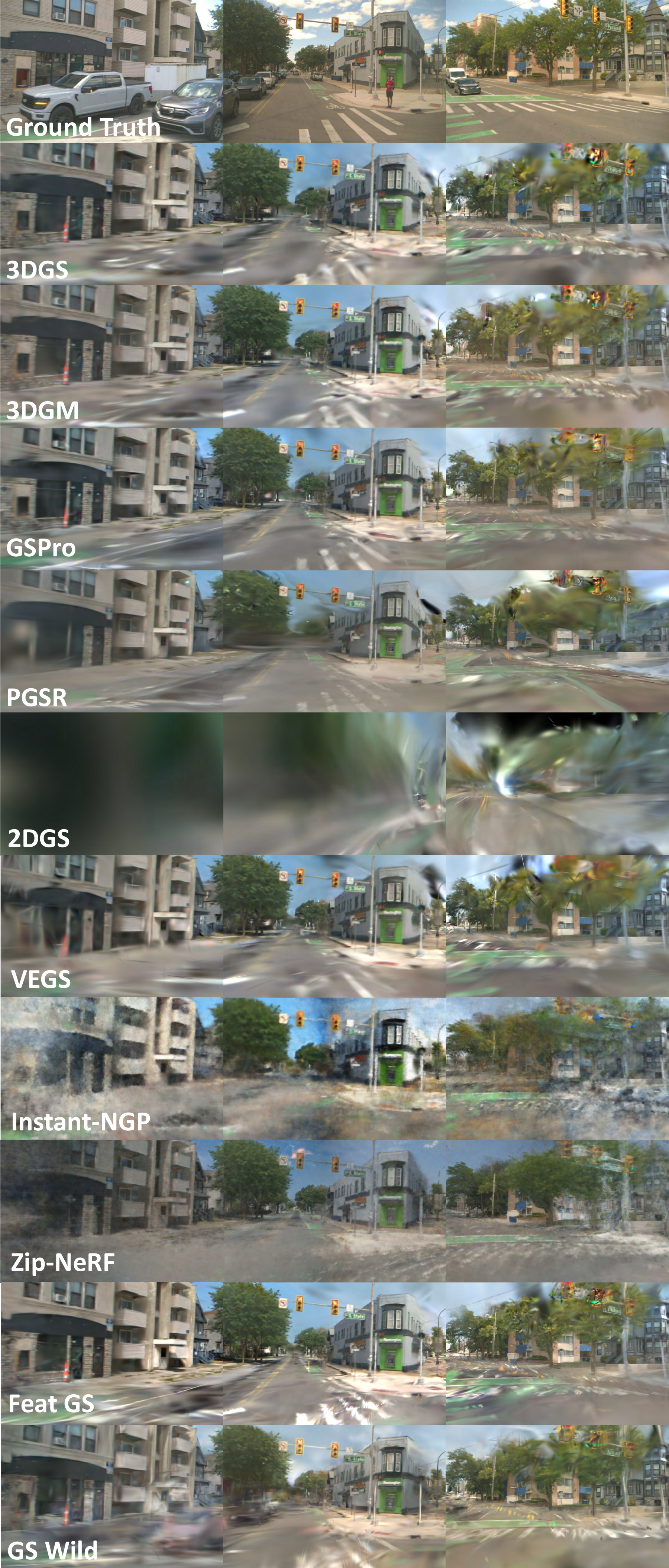}
        \caption{Rendering results comparison in extrapolated view.}
        \label{fig: Detailed results setting 3 test}
    \end{subfigure}
    \caption{\textbf{Qualitative comparison of baseline methods in Setting 3.} The rendering quality deteriorates significantly in extrapolated viewpoints. The geometry becomes fragmented, especially in trees, traffic lights, and lane marks.}
    \label{fig:Qualitative results comparison across all baselines at setting 3}
\end{figure*}

\noindent \textbf{Qualitative Comparison.} We present the qualitative baseline comparison across all settings and baselines in \autoref{fig:Qualitative results comparison across all baselines at setting 1}, \autoref{fig:Qualitative results comparison across all baselines at setting 2} and \autoref{fig:Qualitative results comparison across all baselines at setting 3}. \textbf{(1)} In Setting 1, as shown in \autoref{fig: Detailed results setting 1 test}, all methods exhibit imperfections in ground rendering, while planar-based methods such as 2DGS~\cite{huang20242d} and PGSR~\cite{chen2024pgsr} show comparatively fewer flaws on the ground surface. GSPro~\cite{cheng2024gaussianpro} produces more accurate geometry reconstruction, achieving realistic surfaces and high-fidelity representations of street objects like trees and buildings. \textbf{(2)} In Setting 2, as shown in \autoref{fig:Qualitative results comparison across all baselines at setting 2}, most baselines suffer from sky artifacts such as holes and floating objects. In contrast, VEGS~\cite{hwang2024vegs} produces the more accurate renderings, exhibiting minimal floating artifacts and broken geometry, attributed to the guidance provided by diffusion priors. \textbf{(3)} In Setting 3, as shown in \autoref{fig: Detailed results setting 3 test}, all baselines face significant challenges on the test set. The geometry across all methods appears highly fragmented, and the color consistency is compromised, reflecting a tendency to overfit to the training views. Among the baselines, 2DGS and PGSR show relatively weaker performance, underscoring the limitations of planar representations in effectively capturing the complexity of whole urban scenes.

